%% file: MAIN.tex
\documentclass[3p,twocolumn]{elsarticle}
\usepackage[T1]{fontenc}
\usepackage{booktabs}
\usepackage{CJKutf8}
\usepackage{hyperref}
\usepackage{amsmath}
\usepackage{mathtools}
\usepackage{subfig}
\usepackage{csquotes}
\usepackage{color,soul}
\usepackage{titlesec}
\usepackage[shortlabels]{enumitem}
\usepackage{multirow}
\usepackage{amsfonts}
\usepackage{amssymb}
\usepackage{tikz}
\usepackage{tabularx}
\usepackage{adjustbox}
\usepackage{url}
\usepackage{makecell}
\usepackage{bm}
\usepackage[normalem]{ulem}
\usepackage{floatrow}
\usepackage{todonotes}
\usepackage{CJKutf8}

\journal{ISPRS Journal of Photogrammetry and Remote Sensing}

\begin{document}
\sloppy
\begin{frontmatter}

\title{Building Floor Number Estimation from Crowdsourced Street-Level Images: \\
Munich Dataset and Baseline Method
} 

\author{Yao Sun\corref{mycorrespondingauthor}}
\cortext[mycorrespondingauthor]{Corresponding author}
\ead{yao.sun@tum.de}

\author{Sining Chen}
\ead{sining.chen@tum.de}

\author{Yifan Tian}
\ead{yifan.tian@tum.de}

\author{Xiao Xiang Zhu}
\ead{xiaoxiang.zhu@tum.de}

\address{Data Science in Earth Observation, 
Technical University of Munich, Arcisstraße 21, 80333 Munich, Germany}

\begin{abstract}
\textcolor{blue}{This is the pre-print version. }
Accurate information on the number of building floors, or above-ground storeys, is essential for household estimation, utility provision, risk assessment, evacuation planning, and energy modeling. Yet large-scale floor-count data are rarely available in cadastral and 3-D city databases. 
This study proposes an end-to-end deep-learning framework that infers floor numbers directly from unrestricted, crowdsourced street-level imagery, avoiding hand-crafted features and generalizing across diverse façade styles. 
To enable benchmarking, we release the Munich Building Floor Dataset, a public set of 6 800+ geo-tagged images collected from Mapillary and targeted field photography, each paired with a verified storey label. 
On this dataset the proposed classification–regression network attains 81.2\% exact accuracy and predicts 97.9\% of buildings within ±1 floor. 
The method and dataset together offer a scalable route to enrich 3-D city models with vertical information and lay a foundation for future work in urban informatics, remote sensing, and geographic information science. Source code and data will be released under an open licence at \url{https://github.com/ya0-sun/Munich-SVI-Floor-Benchmark}. 
\end{abstract}


\end{frontmatter}


\begin{CJK*}{UTF8}{gbsn}

\input{sections/01_intro}

\input{sections/02_data-source}
\input{sections/03_dataset}
\input{sections/04_experiments}
\input{sections/05_discussion}
\input{sections/06_conlusion}

\section*{Acknowledgment}

The work is supported by the Deutsche Forschungsgemeinschaft (DFG, German Research Foundation) – 499168241 for the project OpenStreetMap Boosting using Simulation-Based Remote Sensing Data Fusion (Acronym: OSMSim) 
and by the Technical University of Munich (TUM) Georg Nemetschek Institute under the project Artificial Intelligence for the automated creation of multi-scale digital twins of the built world (Acronym: AI4TWINNING). 

\end{CJK*}

\bibliographystyle{elsarticle-num}
\bibliography{mybibfile}

\end{document}

%% file: sections/01_intro.tex
\section{Introduction}\label{sec:intro}

Building floor numbers, i.e., above-ground stories, are crucial for urban planning, disaster management, and resource allocation. 
Floor numbers facilitate household number estimation, aiding in the equitable distribution of utilities such as water, electricity, and waste management services. Additionally, accurate floor number enhances risk exposure assessments by detailing the vertical distribution of occupants and assets, essential for effective disaster mitigation and evacuation planning.  
They also provide energy consumption estimates, providing insights into a building's energy demand based on floor count and usage patterns~\cite{agugiaro2016energy, lwin2009gis, li2024review}. 

\begin{figure*}[!h]
    \centering
    \subfloat[]{
    \includegraphics[height=.173\linewidth]{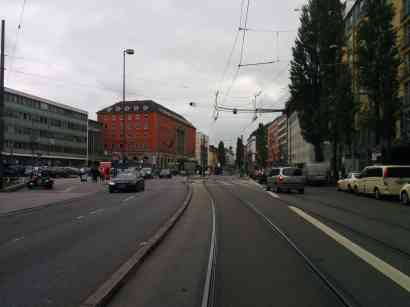}}
    \subfloat[]{
    \includegraphics[height=.173\linewidth]{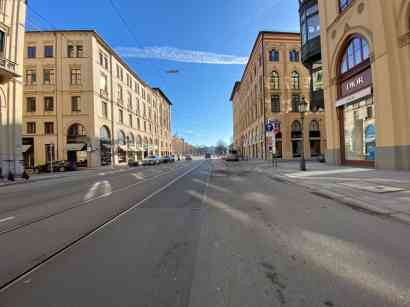}}
    \subfloat[]{
    \includegraphics[height=.173\linewidth]{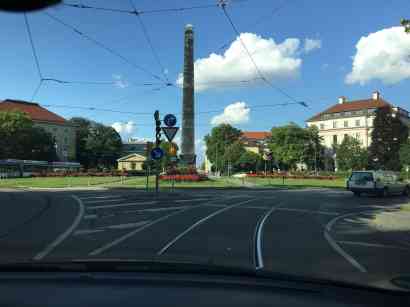}}
    \subfloat[]{
    \includegraphics[height=.173\linewidth]{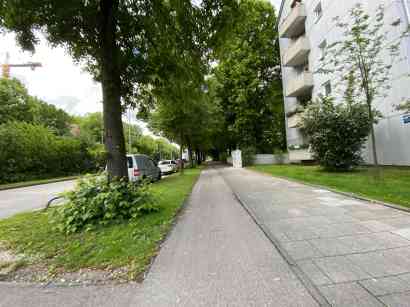}}
    \\
    \subfloat[]{
    \includegraphics[height=.173\linewidth]{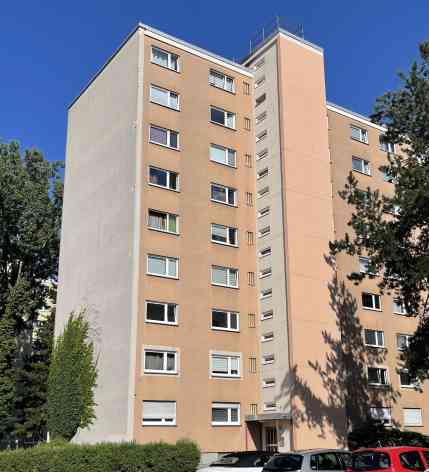}}
    \subfloat[]{
    \includegraphics[height=.173\linewidth]{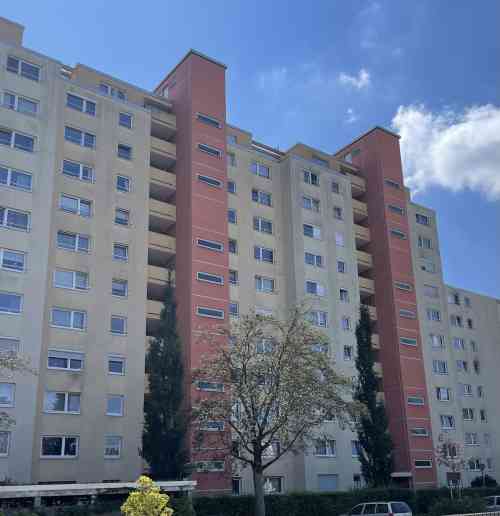}}
    \subfloat[]{
    \includegraphics[height=.173\linewidth]{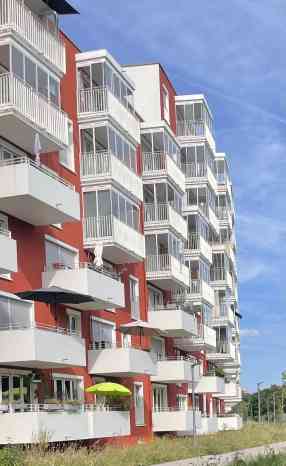}}
    \subfloat[]{
    \includegraphics[height=.173\linewidth]{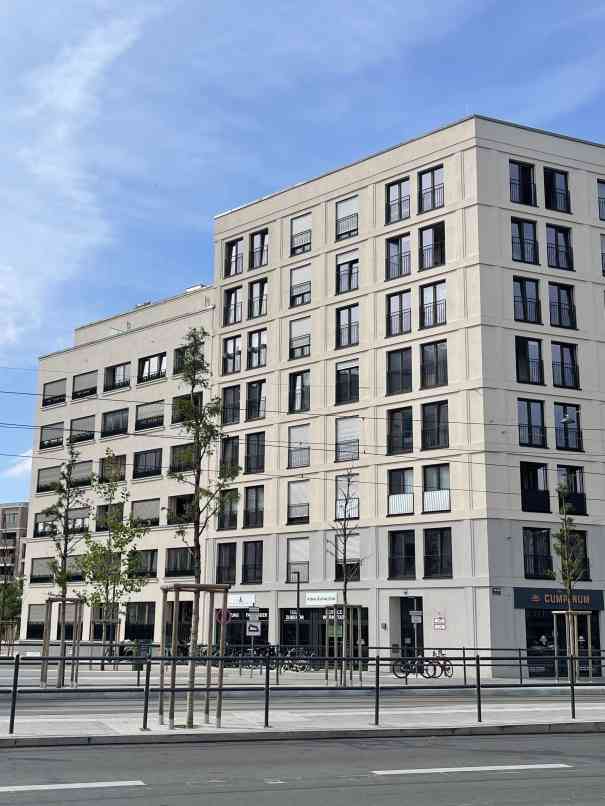}}
    \subfloat[]{
    \includegraphics[height=.173\linewidth]{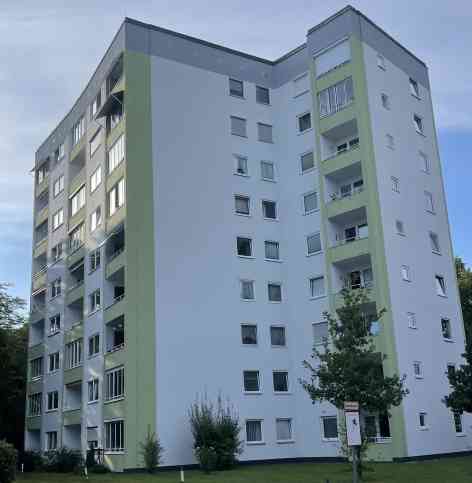}}
    \subfloat[]{
    \includegraphics[height=.173\linewidth]{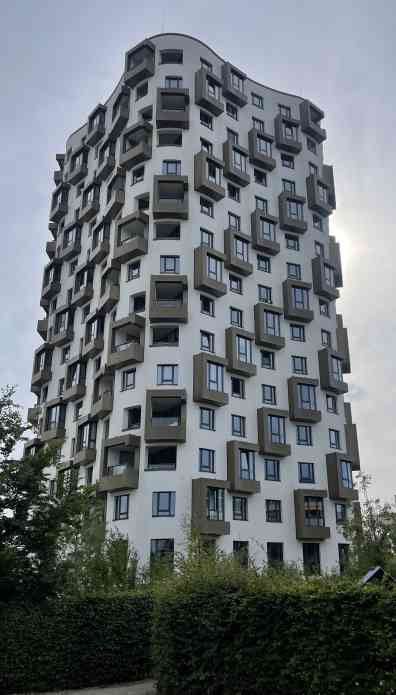}}
    \caption{Examples of SLI used in this study. (a)-(d) Images from Mapillary, showcasing diverse building facades captured from crowdsourced imagery, requiring cropping to isolate buildings and filtering to remove low-quality building images. (e)-(j) Self-captured images, focusing on underrepresented high-rise buildings, with centered framing for clear visualization of facades. }
    \label{fig:photos}
\end{figure*}

Despite its importance, floor number information is often missing in building databases, 
such as cadastral data. Even combined with 3D information, e.g., 3DBAG - 3D building models at multiple levels of detail of the Netherlands~\cite{kadaster_bag,peters2022}, these databases rarely include floor numbers. 
A closely related information is building height, which has received considerable attention in remote sensing research~\cite{zhu2020so2sat, sun2023lcz}. Height estimation methods often employ high-resolution optical imagery~\cite{5653386,chen2021maskH,li20233dcentripetalnet}, LiDAR~\cite{wu2019city}, or synthetic aperture radar (SAR)~\cite{sun2017Building,sun2020cgnet, sun2021bbox,li2023semi} using varies approaches to infer height from features such as shadows~\cite{lee2013automatic}, multi-view geometry~\cite{sirmacek2012performance}, or radar backscatter~\cite{guida2010height}. 
Intuitively, building height and the floor number are correlated; however, their relationship is not linear. 
Assumptions of using a fixed floor height to convert the building height to floor number, such as dividing building height by 3 meters to estimate floor numbers, are invalid due to variability in ceiling heights, roof structures, and construction practices~\cite{Roy2023floor, iannelli2017extensive}. For example, the ground floor often has a different height than other floors, attics may not count as floors but contribute to height, while partial basements, common in European buildings, are excluded from above-ground story counts but included in height measurements. These complexities render height-based approaches unreliable for estimating floor numbers, as noted in studies like~\cite{biljecki2017generating}. 
For building floor number information, manual surveys and field measurements are labor-intensive, time-consuming, and infeasible at scale, underscoring the need for automated approaches~\cite{li2020instance}.

A few studies have explored the automated estimation of building floor numbers. Roy \textit{et al.}\cite{Roy2023floor} inferred floor numbers using multiple building attributes like internal layouts and floor plans, but such attributes are unavailable for most buildings, limiting scalability. 
In contrast, street-level imagery (SLI)\footnote{
\emph{Street-level imagery} (SLI) denotes ground-based photograph taken from pedestrian or vehicle height.  \emph{Street-view imagery} (SVI) is considered as a subset of SLI, where the road is typically centered. The two terms are often used interchangeably. 
For brevity, we use the abbreviation {SLI} throughout this paper.
}
 offers a practical data source for floor number estimation due to its detailed depiction of building facades, capturing visual cues such as windows and other architectural features~\cite{sun2023flickrstr, tian2024learning}. Google Street View (GSV), in particular, has been widely utilized for such tasks because of its high-quality panoramic views and extensive coverage. 
 Iannelli and Dell'Acqua~\cite{iannelli2017extensive} classified building stories in San Francisco into five categories, ranging from no buildings (0) to four or more floors (4+), using GSV images. However, reliance on GSV poses challenges due to restrictive licensing, limiting large-scale dataset creation and algorithm deployment. 
 Crowdsourced platforms like Mapillary, KartaView, and Flickr offer an alternative with open licensing, diverse image contributions, and growing data coverage~\cite{sun2023osmstr}. 
Images from platforms often lack panoramic coverage or calibrated geo-positioning, leading to missing building tops and suffer from less accurate metadata and lower image quality compared to GSV~\cite{sun2023flickrstr}.  
Despite these limitations, the diverse, real-world conditions captured in crowdsourced images enhance their usefulness for developing robust, scalable algorithms. 

Related to building floors, Wu \textit{et al.}~\cite{wu2021floorlevel} proposed FloorLevel-Net, a multi-task learning network that segments facades into semantic regions and floor-level distributions, refining outputs into polylines with geometric parameters to support augmented reality applications. Some studies have explored clustering approaches to detect and parse windows for floor estimation, demonstrating the potential of structured analysis of facade features~\cite{dobson2023floor, kramm2023detecting}, showing the feasibility of using facade features for floor estimation. 
However, vertical window counting is hindered by irregular facade patterns and roof structures, which introduce ambiguity. These limitations underscore the need for scalable, flexible data sources and methods better suited to diverse urban environments.

This work addresses the limitations of existing methods by leveraging unrestricted, crowdsourced SLI to develop scalable and automated approaches for estimating building floor numbers. 
We propose an end-to-end deep learning framework that eliminates the reliance on manually engineered features, thereby enhancing adaptability across diverse building styles. Furthermore, we introduce an open-access dataset for the city of Munich, a heterogeneous collection of images and aims to address challenges related to scalability and data availability.  

Our contributions are threefold:
\begin{itemize}
    \item We developed an automated pipeline to extract building images from Mapillary data and match them to corresponding building footprints.
    \item We created a diverse dataset of over 6.8k SLIs with annotated floor numbers, combining both publicly available Mapillary imagery and self-captured photographs from Munich. 
    \item We introduce a classification-regression network as a baseline for floor number estimation, achieving an accuracy of 81.19\% and 97.90\% within a one-floor error margin.

\end{itemize}

 The remainder of this paper proceeds as follows. 
 Section~\ref{sec:data_source} introduces the study area and data sources, and 
 Section~\ref{sec:data_gen} is concerned with the dataset generation approach to tackle the problem of dataset scarcity. 
 Section~\ref{sec:method} formulates the problem and delineates the proposed method.  
 The experiments and results are presented and analyzed in Section~\ref{sec:test}. 
 In Section~\ref{sec:dis}, we discuss several practical problems related to applying our method to large-scale building floor number retrieval. 
 Section~\ref{sec:conclude} concludes this paper.

%% file: sections/02_data-source.tex
\section{Study Area and Data Sources}\label{sec:data_source}

\subsection{Study Area}\label{sec:studyArea}

\begin{figure*}[]
    \centering
    \subfloat[ ]{\includegraphics[width=.495\textwidth, height=.25\textwidth]{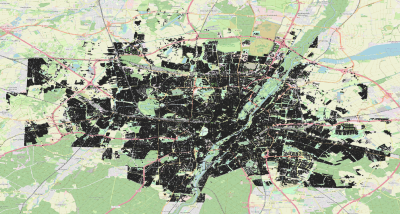}}\hfill
    \subfloat[]{\includegraphics[width = .495\textwidth,height=.25\textwidth]{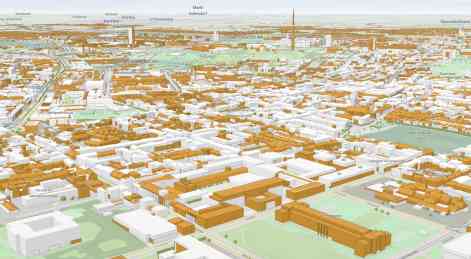}}
    \caption{(a) Coverage of LoD2 building models in Munich, and (b) Example of LoD2 building models in Munich. }
    \label{fig:Lod2_muc}
\end{figure*}

Our study area is Munich, Germany, a mid-size European city whose architectural styles (historic low-rise blocks and modern towers) typify many European urban areas. 
The study area is defined by a bounding box corresponding to the extent of the Level of Detail 2 (LoD2) dataset coverage~\cite{BayerischeVermessungsverwaltung2021} of of Munich (c.f., Fig.~\ref{fig:Lod2_muc} (a)), whose south-west and north-east corners lie at (11.3212° E, 48.0557° N) and (11.7774° E, 48.2872° N), respectively. 
This spatial extent sets the stage for acquiring facade imagery, described in Section~\ref{sec:data_gen}.

\subsection{Street-level Images (SLIs)}

To obtain facade views, we compiled SLIs from two complementary sources: (i) crowdsourced photos retrieved from Mapillary, and (ii) photographs we captured ourselves, as shown in Fig.~\ref{fig:photos}. These sources differ in coverage, framing, and quality, complementing each other to enhance diversity and ensure representation of various building heights and styles. 
Below, we introduce these data sources, and the processing pipeline is detailed in Section~\ref{sec:data_gen}.

\begin{itemize}
    \item {Mapillary SLIs}

Mapillary images were selected for their global coverage, open access API, and detailed metadata, including capture time, coordinates, camera type, and quality score. SLIs within the Munich boundary were retrieved and filtered to exclude low-quality and nighttime images. 
To retrieve building images, the original images were processed using the Grounding DINO model~\cite{liu2023grounding} to detect and crop individual buildings. 

These images often place streets at the center with buildings on either side, requiring cropping to isolate facades. The data offers extensive coverage and diversity; while challenges include occlusions from vegetation or vehicles, incomplete building tops or bases, particularly for high-rise structures, and therefore the need to filter out a large number of unsuitable images.

    \item Self-captured SLIs

To address gaps in Mapillary's coverage, particularly for high-rise buildings, we manually captured additional images. These self-captured images ensured that buildings were centrally framed, reducing the need for detection and cropping. 

Although this dataset is smaller in scale and coverage, it complements the Mapillary images by adding underrepresented building types. Challenges were less frequent, with fewer issues related to occlusions or incomplete structures, and a smaller proportion of images requiring filtering.

\end{itemize}

\subsection{Ground Truth of Building Floor Numbers and Footprints}\label{sec:GT}

Ground-truth floor counts and footprints are sourced from the Munich LoD2 CityGML models, which provide detailed 3D city models compliant with the ALKIS (Authoritative Real Estate Cadastre Information System)\cite{BayerischeVermessungsverwaltung2021}. These models include attributes such as building height, roof type, ground elevation, story count, and building function, with simplified roof structures as defined by CityGML LoD2 standards~\cite{open2012ogc}.

\begin{figure*}
    \centering
    \includegraphics[width=.8\linewidth,trim=0cm 0cm 0cm 0.73cm, clip]{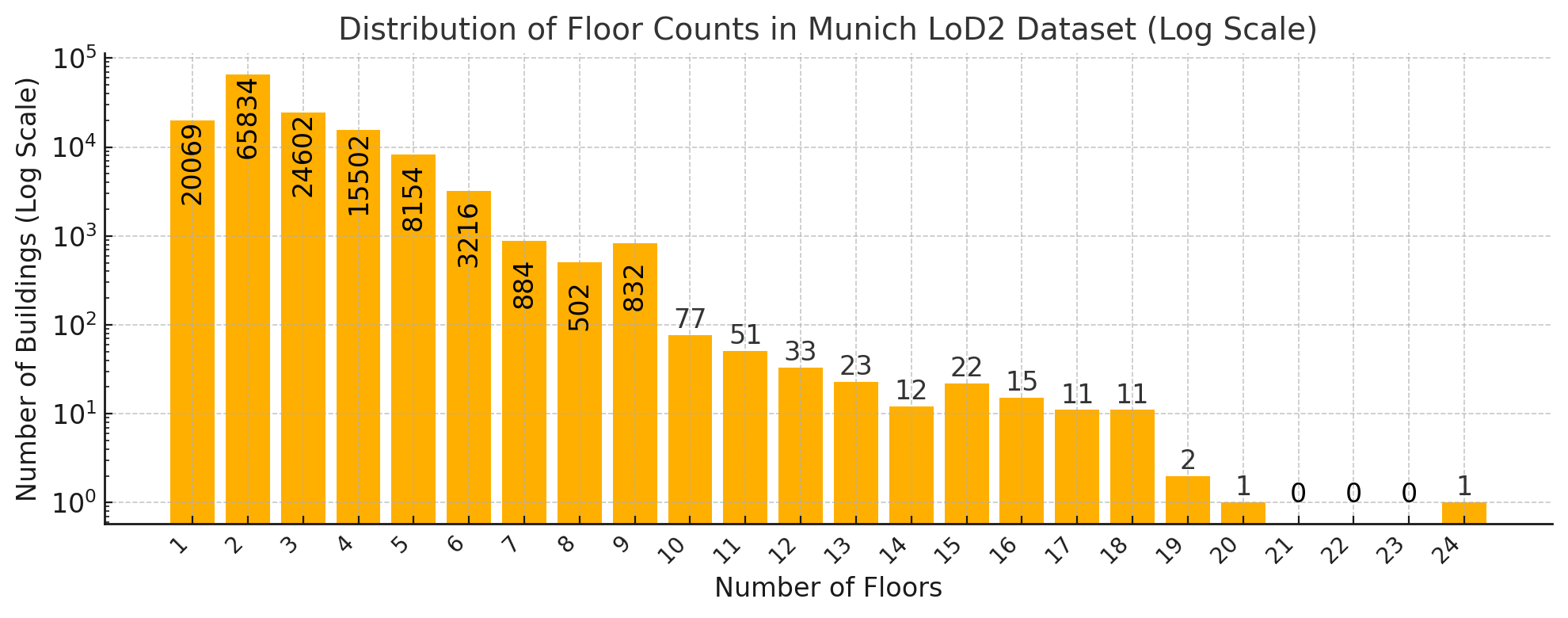}
    \caption{Distribution of Floor Counts in Munich LoD2 CityGML Database (Log Scale), with building counts labeled on each bar. }
    \label{fig:lod-floorCounts}
\end{figure*}

The Munich LoD2 dataset contains 372,120 buildings, with approximately one-third (139,854) having above-ground story data. As shown in Fig.~\ref{fig:lod-floorCounts}, 
the floor count distribution reveals that most buildings (over 10,000) have 1 to 4 floors, with 2 floors dominating (65,834). Mid-rise classes Buildings with 5 to 6 floors occur in thousands, while those with 7 to 9 floors number between 500 and 1,000. high-rise structures (10 or more floors) are sparse, only four exceed 18 floors. 

\begin{figure*}[h!]
    \centering
    \subfloat[Residential buildings]{\includegraphics[width=.47\linewidth]{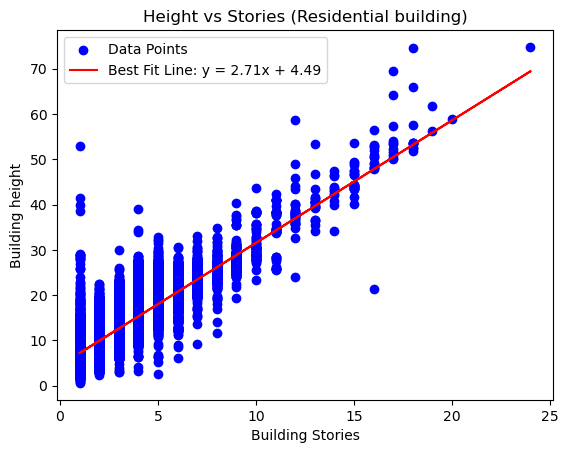}\label{fig:4_residential}}
    \hfill
    \subfloat[Commercial buildings]{\includegraphics[width=.47\linewidth]{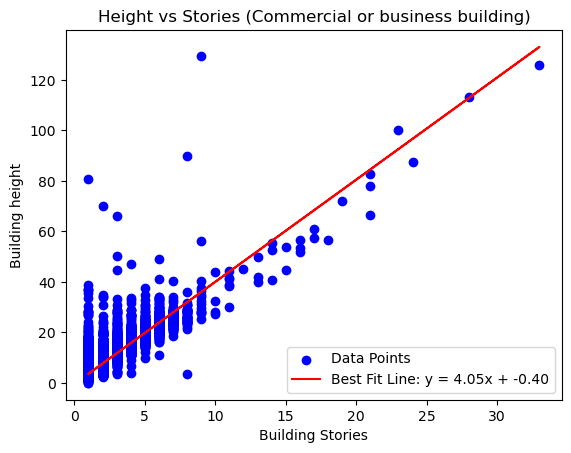}\label{fig:4_business}}
    \\
    \subfloat[Flat roofs]{\includegraphics[width=.47\linewidth]{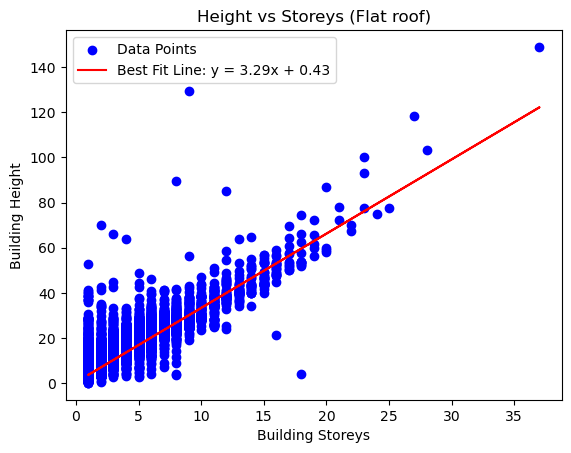}\label{fig:4_flat}}
    \hfill
    \subfloat[Hipped roofs]{\includegraphics[width=.47\linewidth]{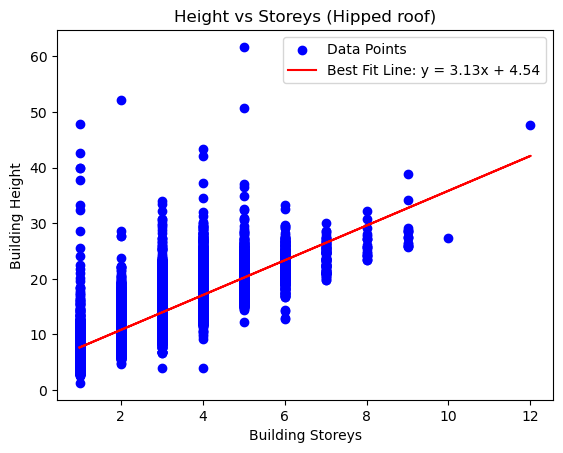}\label{fig:4_hipped}}
    \caption{Scatter plots of building floor and height for selected building functions and roof types.}
    \label{fig:4_scatter_combined}
\end{figure*}

We analyzed the relationship between building height and floor count. 
While a general trend of increasing height with more stories is observed, significant variability exists within each category, indicating no linear relationship. This is true even for buildings with similar usage or roof types, such as residential or flat-roofed buildings, showing in Fig.~\ref{fig:4_scatter_combined}. Factors like roof type, building function, architectural design, and construction era contribute to this variability, highlighting the need for estimation methods that go beyond simple height-based assumptions.

%% file: sections/03_dataset.tex
\section{Dataset Generation}\label{sec:data_gen}

This section describes the pipeline for constructing the dataset, from image acquisition to quality control. We rely on two complementary SLI sources: crowdsourced Mapillary photos and self-captured images, as detailed below.

\subsection{Crowdsourced Mapillary Data}

We first acquired Mapillary images covering Munich. 
The images then pass through four steps: (i) retrieval and filtering; (ii) facade detection and cropping; (iii) image–footprint matching; (iv) quality control and augmentation.

\subsubsection{Retrieval and Filtering} 

We queried the Mapillary API for all photos within the Munich bounding box (c.f., Section~\ref{sec:studyArea}). Metadata (time, pose, camera model) enabled automatic exclusion of panoramas, night shots and low-quality images. Processed metadata and image IDs were stored in JSON and GPX formats for subsequent steps. 

\subsubsection{Building Detection and Cropping}

\begin{figure}[!]
    \centering
    \subfloat[Detection example]{
    \includegraphics[height=3cm]{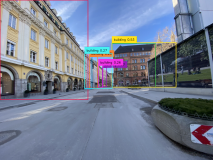}
    \label{fig:detection_ex1}
    }
    \subfloat[Cropped building]{
    \includegraphics[height=3cm]{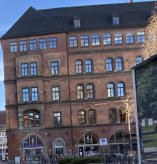}
    \label{fig:cropping_ex1}
    }
    \caption{Examples of building detection (a) and cropping (b).}
    \label{fig:detect_and_crop}
\end{figure}

The Grounding DINO model~\cite{liu2023grounding} was used to detect buildings within SVIs by generating bounding boxes to differentiate individual structures (Fig.~\ref{fig:detect_and_crop}). 
The model, initialized with a "building" textual prompt, applied thresholds for box detection and text relevance. Bounding boxes below a specified width were excluded to maintain clarity. The resulting bounding box data were stored in JSON format for subsequent cropping and matching steps. Using this information, buildings were cropped from the SVIs (such as shown in Fig.~\ref{fig:cropping_ex1}), streamlining integration into the analytical pipeline.

\subsubsection{Image-to-Footprint Matching}

\begin{figure}[!]
    \centering
    \includegraphics[width=0.5\linewidth]{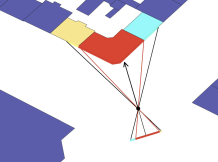}
    \caption{Schematic representation of the matching algorithm.}
    \label{fig:matching_principle}
\end{figure}

Matching SVIs to corresponding building footprints from the Munich LoD2 dataset was conducted through a multi-step process. Circular buffer zones were created around each SVI position to clip building footprints, focusing on relevant areas and reducing computational load. The field of view (FoV) for each cropped image was calculated using bounding box positions and camera orientation to ensure accurate alignment. Rays were projected from the camera position within the FoV to identify intersected footprints, with matches determined by the frequency of intersected footprint IDs (Fig.~\ref{fig:matching_principle}). A binary search strategy refined the process, improving efficiency by minimizing the number of rays required for precise identification \cite{open2012ogc, geopy2024}.

\subsubsection{Quality Control and Augmentation }

\begin{figure}[!]
    \centering
    \subfloat{
        \includegraphics[height=2.6cm]{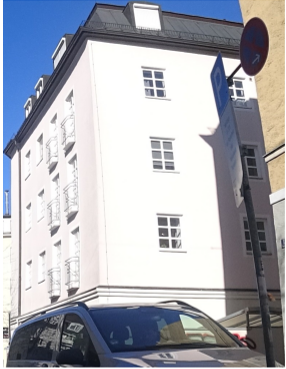}
        \label{fig:SVI_9319}}   
    \subfloat{
    \includegraphics[height=2.6cm]{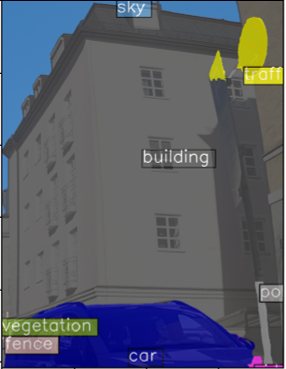}}
    \hfill
    \subfloat{
        \includegraphics[height=2.6cm]{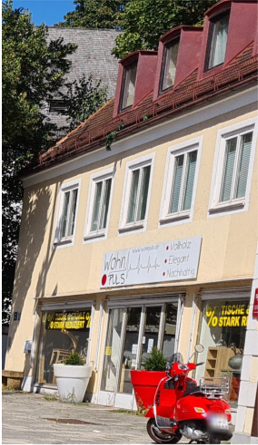}
        \label{fig:SVI_9319}}  
    \subfloat{
    \includegraphics[height=2.6cm]{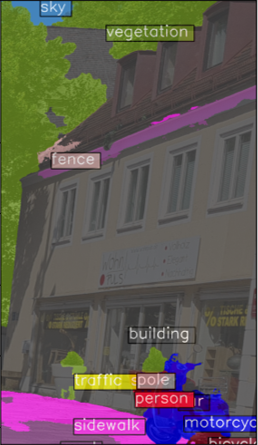}}
    \caption{Examples of semantic segmentation results.} 
    \label{fig:SVI_seg}
\end{figure}

Semantic segmentation was performed using the MMSegmentation framework \cite{contributors2020mmsegmentation} to label building images and identify obstructions such as vegetation and sky. The segmentation results were used to calculate pixel proportions for each category, establishing thresholds for filtering images. Images were excluded if the proportion of building pixels was low or vegetation pixels was high (thresholds empirically set to 20\% and 70\%, respectively). 
Images were also excluded if windows were not detected using the Grounding DINO model~\cite{liu2023grounding}. Additionally, images where building segments dominated the top rows or were obscured by vegetation at the top were also removed based on specific thresholds. Fig.~\ref{fig:SVI_seg} illustrates examples of the segmentation results.

\begin{figure}[!]
    \centering
    \subfloat{
        \includegraphics[height=2cm]{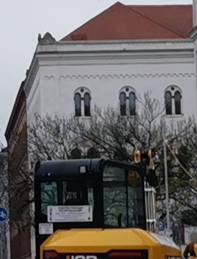}
        \label{b1}}
    \hfill
    \subfloat{
        \includegraphics[height=2cm]{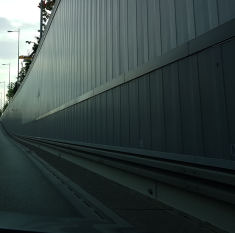}
        \label{b2}}
    \hfill
    \subfloat{
        \includegraphics[height=2cm]{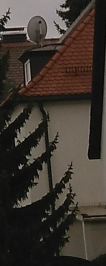}
        \label{b3}}
    \hfill
    \subfloat{
        \includegraphics[height=2cm]{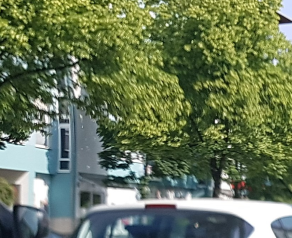}
        \label{b4}}
    \caption{Examples of building images removed during manual inspection.}
    \label{fig:4_badcase}
\end{figure}

Finally, images were manually reviewed to ensure the dataset contained only high-quality images suitable for training. Poor-quality images, such as those with incomplete buildings, excessive obstructions, or insufficient structural detail, were removed (Fig.~\ref{fig:4_badcase}). After determining usable building images, they are rectified and augmented for the next stage. 

The final Mapillary subset contained 6,473 building images. These images were rectified and augmented to ensure suitability for model training. 

\subsection{Supplementary Self-Captured Imagery} 

Mapillary contained few examples of tall buildings (c.f., Section~\ref{sec:GT}), After analyzing Munich's building distribution and the limitations of the initial dataset, 
To balance the long tail, we photographed additional high-rise buildings. 

As shown in Fig.~\ref{fig:lod-floorCounts}, 
Munich's building distribution includes a comparable number of 7–9 floor buildings (500–900 each) and fewer buildings with 10–18 floors, and 
buildings with more than 18 floors are extremely rare in Munich (only four instances records in the Munich LoD2 database). 
Given their scarcity, we excluded these buildings from the dataset to maintain focus on categories with sufficient representation. 
Therefore, 
self-captured images are planned to cover buildings with 8–18 floors to improve the representation of high-rise buildings. This practical and targeted approach provided flexibility to address specific gaps in the dataset. 

\subsubsection{Building Selection} 
Based on the floor count distribution of residential buildings in Munich, we determined the required number of images for each floor category. Buildings to be photographed were then randomly selected from the LoD2 database to ensure broad geographic dispersion across the city. 

\subsubsection{Route Optimisation}
\begin{figure}
    \centering
    \includegraphics[width=\linewidth]{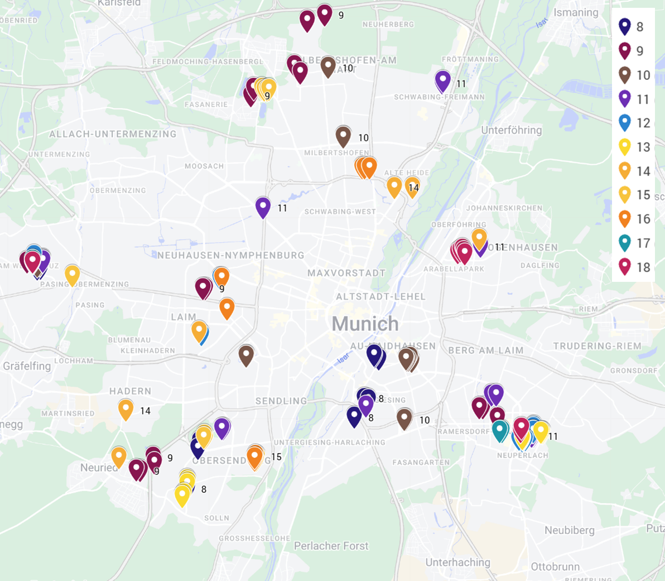}
    \caption{The geographic distribution of self-captured buildings in Munich, color-coded by floor number.}
    \label{fig:photo-shoot}
\end{figure}

To ensure efficient image acquisition, we planned a route to reach all target buildings. First, we calculated the centroid coordinates of building polygons as target points. Using these coordinates, we solved the Traveling Salesperson Problem (TSP) to determine the optimal visiting sequence by minimizing travel distances. A distance matrix was generated for all building pairs, and the TSP solution produced an ordered list of building points. 
We then generated a map using Google Maps to visualize and guide the route for capturing images. Fig.~\ref{fig:photo-shoot} shows the geo-distribution of the self captured data, and the map and the data table are made publicly accessible \href{https://www.google.com/maps/d/u/0/edit?mid=1I4hB6QzSnvfQcbCEAprPWpDaNQVg52g&usp=sharing&hl=en}{at this link}.

\subsubsection{Capture Protocol}
Following the planned route, we captured three images (N=3) of each selected building from different perspectives to enhance variability and coverage. The number of images was fixed for each floor category to ensure consistency across the dataset. 

\subsection{Dataset Overview and Statistics}

The resulting dataset comprises a total of 6,827 building images with ground truth data on the floor number. 
Among these, 6,473 images were sourced from Mapillary, and 354 images were self-captured particularly for buildings with more than seven floors. The detailed distribution of images across different story numbers is shown in Table~\ref{tab:building_story_dataset}.

\begin{figure}
    \centering
    \includegraphics[width=\linewidth,trim=0.28cm 0cm 0.4cm 0.71cm, clip]{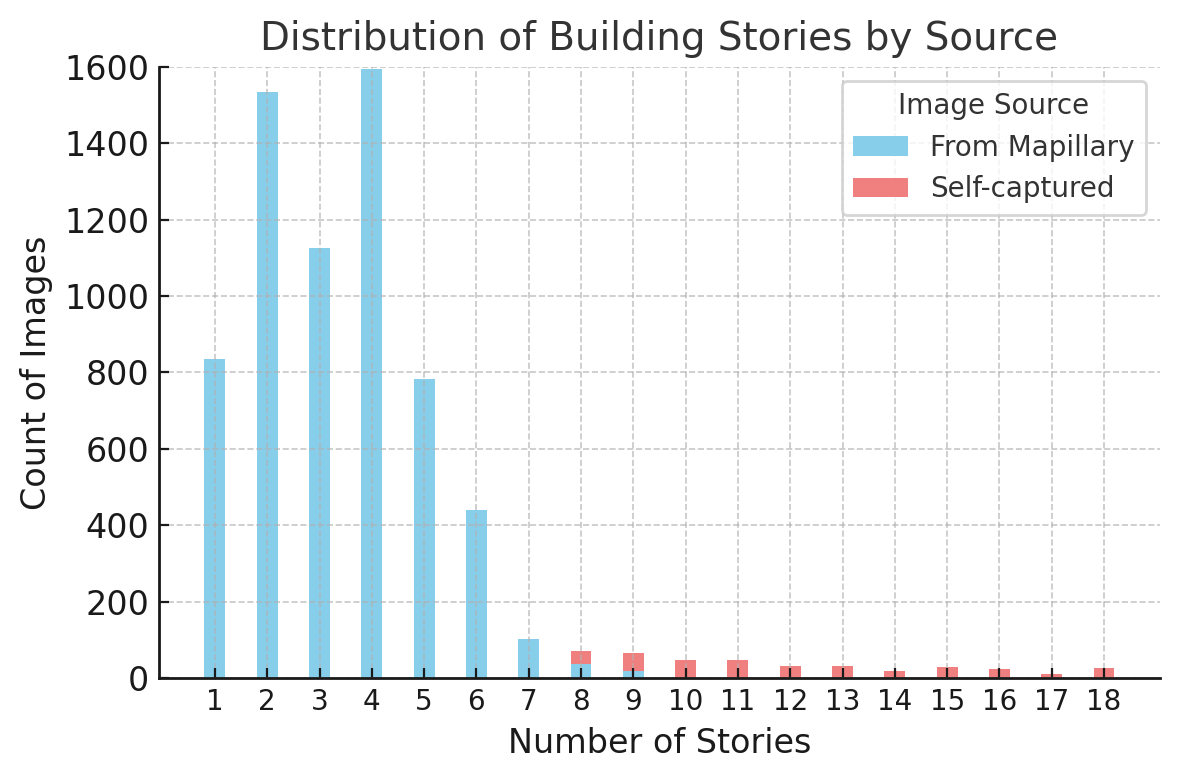}
    \caption{Distribution Of Building Stories By Source.}
    \label{fig:hist_count}
\end{figure}

Fig.~\ref{fig:hist_count} shows the distribution of building floors by source, exhibiting a pronounced long-tail distribution. 
As can be seen, a high concentration of images for buildings with fewer floors and a gradual decline as the floor number increases. While most images represent low-rise buildings, significant number of images depicting high-rise buildings, though each occurs less frequently. 
Regarding the sources, 
buildings with 1–7 floors from Mapillary, 
10–18 floors from the supplementary self-captured data, 
and 8-9 floors from both sources.

\begin{table}[h!]
\footnotesize
\centering
\caption{Building floor dataset: Total images and sources}
\label{tab:building_story_dataset}
\begin{tabular}{rrrr}
\hline
\textbf{Floor} & \textbf{Total} & {From} & {Self-} \\ 
\textbf{count} & \textbf{Image number} & {Mapillary} & {captured} \\
\hline
1                       & 835                   & 835               & 0                          \\
2                       & 1535                  & 1535              & 0                          \\
3                       & 1126                  & 1126              & 0                          \\
4                       & 1594                  & 1594              & 0                          \\
5                       & 784                   & 784               & 0                          \\
6                       & 441                   & 441               & 0                          \\
7                       & 103                   & 103               & 0                          \\
8                       & 70                    & 37                & 33                         \\
9                       & 66                    & 18                & 48                         \\
10                      & 48                    & 0                 & 48                         \\
11                      & 48                    & 0                 & 48                         \\
12                      & 33                    & 0                 & 33                         \\
13                      & 33                    & 0                 & 33                         \\
14                      & 18                    & 0                 & 18                         \\
15                      & 30                    & 0                 & 30                         \\
16                      & 24                    & 0                 & 24                         \\
17                      & 12                    & 0                 & 12                         \\
18                      & 27                    & 0                 & 27                         \\ \hline
\textbf{Total}          & \textbf{6827}         & {6473}     & {354}               \\ \hline
\end{tabular}
\end{table}

%% file: sections/04_experiments.tex
\begin{figure*}[h!]
    \centering
    \includegraphics[width=\linewidth]{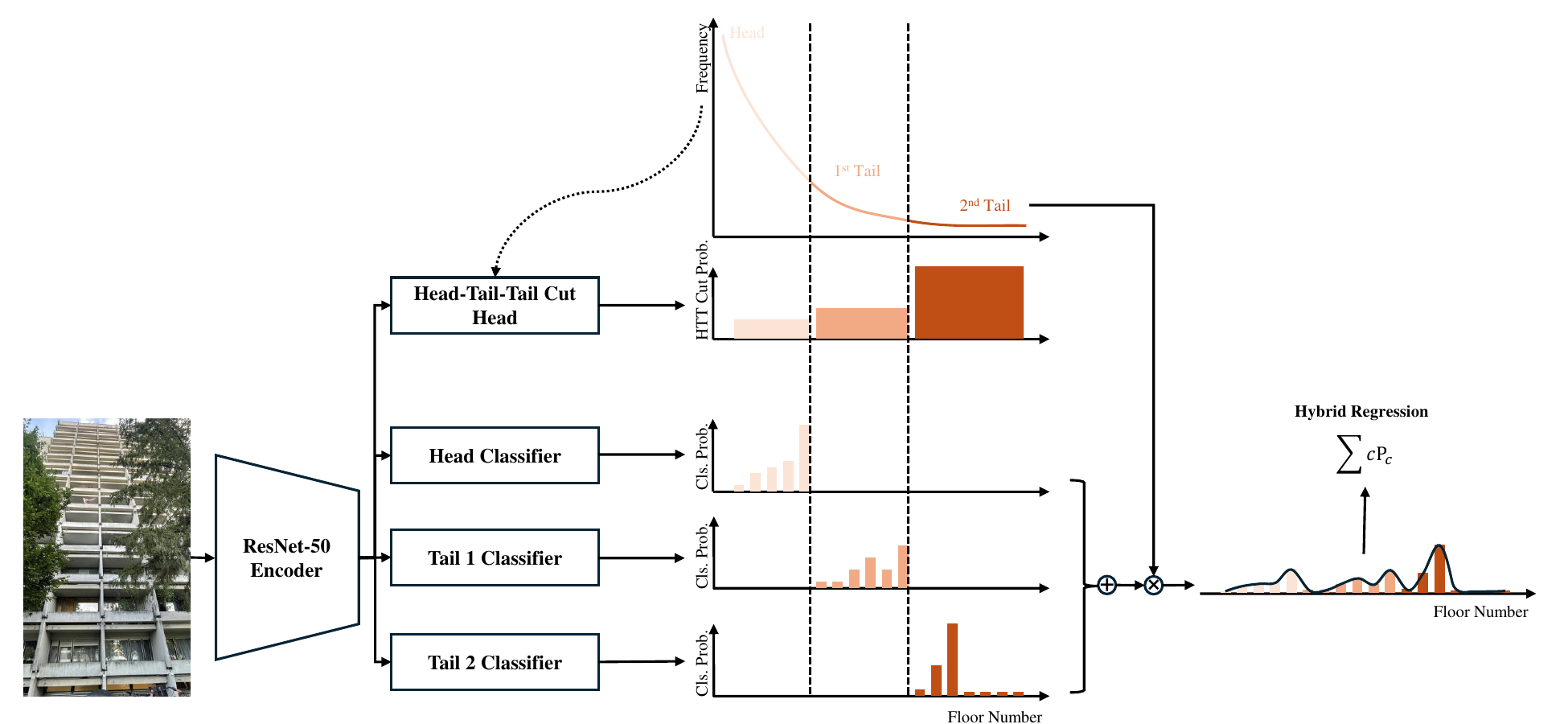}
    \caption{Network Architecture. The network architecture consists of a ResNet-50 \cite{he2016resnet} encoder, a HTTC head, and three classifiers, each corresponding to one of the three cut ranges. The HTTC is designed to divide the dataset into three subsets with more balanced distributions, and it predicts the probability that an image belongs to each of these subsets. Each of the three classifiers predicts the class probabilities for the classes within its respective subset. These class probabilities are weighted by the probabilities from the HTTC and concatenated to form the final class probability distribution. The final prediction is computed as the expectation of this class probability distribution. }
    \label{fig:network}
\end{figure*}

\section{Benchmark Network for Floor Estimation}\label{sec:method}

We formulate the task of floor estimation as an \emph{ordinal classification} task and introduce a deep neural network that serves as a reproducible baseline. 

Floor labels are discrete yet ordered. 
Pure classification ignores ordinality, whereas pure regression ignores class boundaries. We therefore adopt a {hybrid regression} (HYB) strategy that combines both views. 
On the other hand, class imbalance is severe (c.f, Table~\ref{tab:building_story_dataset}); most samples lie in low-rise classes. We mitigate this with a {head–tail–tail cut} (HTTC) that splits the label space into three frequency-balanced subsets. 

In the proposed network, as shown in Fig.~\ref{fig:network}, 
a ResNet-50 encoder~\cite{he2016resnet} first extracts image features. These features feed into the HTTC head, which predicts the probability that a facade belongs to each of the three subsets. The HTTC is intended to divide the whole dataset into several less imbalanced subsets. Based on observations of the dataset's distribution, as shown in Table~\ref{tab:building_story_dataset} and visualized in Fig.~\ref{fig:hist_count}, we set the cut points at Class 5 and Class 11, resulting in three subsets:
\begin{itemize}
    \item Head $\mathcal{H}$, consisting of buildings with fewer than 6 floors.
    \item First tail $\mathcal{T}_1$, consisting of buildings with 6 to 10 floors.
    \item Second tail $\mathcal{T}_2$, consisting of buildings with more than 10 floors.
\end{itemize}

The HTTC head is designed to output the probabilities that an image belongs to each of these subsets. It is trained via a cross entropy loss written as

\begin{equation}
\begin{aligned}
    L_{htt} = & - (c_{GT} \in \mathcal{H}) \log p_h 
                - (c_{GT} \in \mathcal{T}_1) \log p_{t1} \\
              & - (c_{GT} \in \mathcal{T}_2) \log p_{t2},
\end{aligned}
\end{equation}

where $c_{GT}$ is the ground truth floor class (i.e., the floor number minus one), and $p_h$, $p_{t1}$, and $p_{t2}$ are predicted probabilities of an image belonging to the head, the first tail, and the second tail, respectively. The sets $\mathcal{H}$, $\mathcal{T}_1$, and $\mathcal{T}_2$ are defined as
\begin{eqnarray}
    \mathcal{H} &=& \{0,1,2,3,4\},\\
    \mathcal{T}_1 &=& \{5, 6, 7, 8, 9, 10\},\\
    \mathcal{T}_2 &=& \{11, 12, 13, 14, 15, 16, 17\}.
\end{eqnarray}

Corresponding to the HTTC, three classifiers predict the class probabilities within each subset. These class probabilities are then weighted by the probabilities from the HTTC and concatenated to yield the overall class probabilities. The process can be expressed as
\begin{equation}
    P_c = 
    \begin{cases}
        p_h p_c, & \text{if } c\in \mathcal{H}, \\
        p_{t1} p_c, & \text{if } c\in \mathcal{T}_1, \\
        p_{t2} p_c, & \text{if } c\in \mathcal{T}_2,
    \end{cases}
\end{equation}
where $p_c$ represents the class probability for floor class $c$, and $P_c$ denotes the final probability for class $c$. The final probability is then used to compute a cross entropy loss, defined as
\begin{equation}
    L_{ce}=-\log P_{c_{GT}}.
\end{equation}

The final class probabilities form a probability mass function (PMF), and the expectation value of this PMF is taken as the raw output, expressed as
\begin{equation}
    F=\sum_c cP_c.
\end{equation}
The raw output $F$ is then used to compute a regression loss in the form of an L1 loss, defined as:
\begin{equation}
    L_{reg}=|F-c_{GT}|.
\end{equation}

During training, all the heads are trained jointly with a total loss, which is the sum of all individual loss terms, expressed as
\begin{equation}
    L=L_{htt}+L_{ce}+L{reg},
\end{equation}
During inference, the predicted floor number is given by the floor of the raw outputs, denoted as $[F]$.

This proposed network serves as a baseline for the dataset, and is evaluated against several variants in Section~\ref{sec:test}.

\section{Experiments and Results}\label{sec:test}
\subsection{Experimental Setup} \label{sec:train_details}

To evaluate the performance of the proposed network, we define a plain classification network. 
We then performed {\textit{Ablation studies}} to demonstrate the efficacy of the designed components, i.e., HTTC and HYB. 
In addition, to assess the impact of the cut point selection for the HTTC, we conducted an experiment using a single cut point at floor class 6 for the head-tail cut (HTC). 
Thus, \textit{{five}} experiments are conducted: 

\begin{itemize}[leftmargin=0.5cm]
    \item Experiment 1. Baseline: a plain classification network; 

    \item Experiment 2. Baseline+HTC(6): A HTC is applied between classes less than 6 and those greater than or equal to 6; 
    
    \item Experiment 3. Baseline+HTTC(5, 11): A HTTC is applied with cut points at classes less than 5, from 5 to 11, and greater than or equal to 11; 

    \item Experiment 4. Baseline+HYB: HYB, where the average of the floor number classes, weighted by the class probabilities, is taken as the output.

    \item Experiment 5. Ours, Baseline+HYB+HTTC(5, 11): HYB plus a HTTC at Class 5 and 11. 

\end{itemize}

All the networks are trained using the Adam optimizer with a learning rate of 0.001 for 100 epochs, during which the models with the highest validation accuracy are selected for evaluation. 
Each experiment is repeated four times, and we report mean ± standard deviation.

\setlength{\tabcolsep}{5pt}
\begin{table*}[h!]
\footnotesize
    \centering
    \begin{tabular}{cccccc}
    \hline
        Experiment & \textbf{Method} & \textbf{Accuracy} (\%) $\uparrow$ & \textbf{Accuracy (±1)} (\%) $\uparrow$ & \textbf{MAE} (floor) $\downarrow$ & \textbf{RMSE} (floor) $\downarrow$ \\
        \hline
        1 & Baseline & 80.35 ± 0.62 & 97.16 ± 0.61 & 0.2439 ± 0.0111 & 0.6600 ± 0.0751 \\
        2 & +HTC(6) & 76.29 ± 0.23 & 92.01 ± 0.18 & 0.4967 ± 0.1039 & 1.6072 ± 0.3153 \\
        3 & +HTTC(5, 11) & 80.19 ± 0.74 & 97.59 ± 0.17 & 0.2342 ± 0.0079 & 0.5876 ± 0.0251 \\
        4 & +HYB & 80.08 ± 0.10 & 97.57 ± 0.35 & 0.2338 ± 0.0123 & 0.5790 ± 0.0161 \\
        5 & +HYB+HTTC(5, 11) & \textbf{81.19 ± 0.97} & \textbf{97.90 ± 0.24} & \textbf{0.2169 ± 0.0124} & \textbf{0.5428 ± 0.0206} \\
    \hline
    \end{tabular}
    \caption{Experimental results. 
    Each experiment is repeated four times, and the mean values of the metrics along with their standard deviations are reported. 
    The metrics used are Accuracy, Accuracy (±1), MAE, and  RMSE; $\uparrow$: The higher, the better; $\downarrow$: The lower, the better.  
    The best results are highlighted in \textbf{bold}.}
    \label{tab:exp_res}
\end{table*}

\subsection{Evaluation Metrics}

Evaluations are conducted using the following four metrics:
\begin{itemize}
    \item $Accuracy = \frac{\text{TP}}{N}$, where TP represents the number of true positives, and N is the total number of test samples.
    
    \item $Accuracy (±1) = \frac{\text{TP}'}{N}$, where TP\'~denotes the number of true positives within a one-floor error margin, i.e., a prediction is considered to be correct if the difference from the label is at most 1.  
    
    \item Mean Absolute Error (MAE): \\
    $MAE = \frac{1}{N}\sum |[F]-c_{GT}|$.
    
    \item Root Mean Square Error (RMSE): \\
    $RMSE = \sqrt{\frac{1}{N}\sum ([F]-c_{GT})^2}$.
\end{itemize}

The selected evaluation metrics account for both classification and regression aspects of the task. 
Notably, \textit{\textbf{Accuracy (±1)}} 
counts a prediction as correct when it deviates by at most one floor from the ground truth. This tolerance accounts for labeling ambiguity in buildings with non-flat roofs or souterrains. 
Such structures often complicate the exact determination of floor numbers, as partial or non-standard floors may not align with conventional definitions. By incorporating a tolerance, this metric provides a more realistic assessment of model performance, acknowledging these architectural nuances and the variability in labeling practices. 

\subsection{Results}

Table \ref{tab:exp_res} summarises the results of the five variants.  
The plain baseline (Experiment~1) reaches 80.35\% accuracy, while our full model (Experiment~5) raises this to 81.19\% and attains the lowest MAE (0.217 floors).  
Introducing only HYB (Experiment~4) or only HTTC split (Experiment~3) each improves MAE by \(\approx\)4 – 5 \%, but combining both techniques yields the best overall performance.    
Conversely, an ill-placed single cut (Experiment~2) harms performance, underscoring that label-space partitioning must follow the data distribution.  
Standard deviations across four runs remain <1 pp for every metric, indicating stable optimization.

Fig.~\ref{fig:confusion_matrix} shows that most mistakes cluster in adjacent classes, which explains the high Accuracy (±1).  
Relative to the baseline, our model sharpens the diagonal for mid-rise categories (7–10 floors) and reduces systematic under-prediction of the rare high-rise classes (>11 floors).  
Residual errors typically arise when roofs or souterrains are occluded; qualitative examples are given in Fig.~\ref{fig:grid_images} and Fig.~\ref{fig:grid_images_continue}.  

These patterns confirm that treating ordinality (HYB) and imbalance (HTTC) together is key to city-wide floor-count mapping.

%% file: sections/05_discussion.tex
\section{Discussion}\label{sec:dis}

\begin{figure*}[h]
    \centering
    \includegraphics[width=\linewidth] {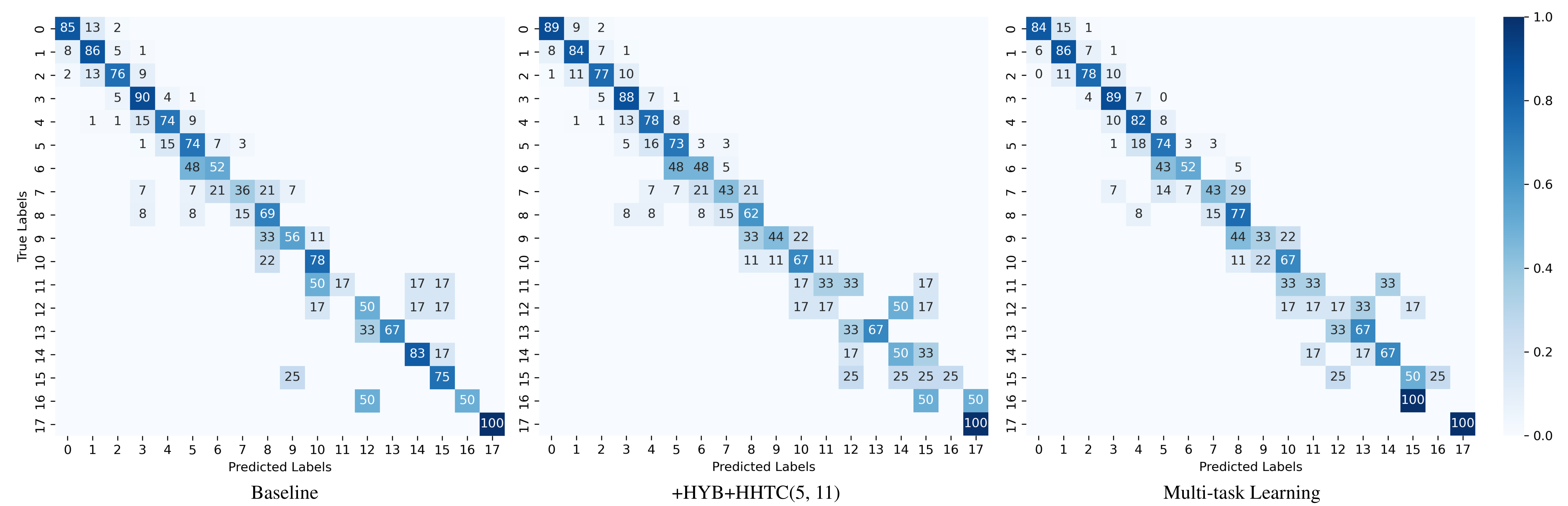}
    \caption{Confusion matrices for the baseline network (Table \ref{tab:exp_res}, Experiment 1), the single-task learning network (Table \ref{tab:exp_res}, Experiment 1 and Table \ref{tab:stl_vs_mtl}, STL), and the multi-task learning network (Table \ref{tab:stl_vs_mtl}, MTL). The values in the matrices represent the percentage of samples for each ground truth label.}
    \label{fig:confusion_matrix}
\end{figure*}

\begin{table*}[h!]
\footnotesize
    \centering
    \begin{tabular}{ccccc}
    \hline
        \textbf{Method} & \textbf{Accuracy} (\%) $\uparrow$ & \textbf{Accuracy (±1)} (\%) $\uparrow$ & \textbf{MAE} (floor) $\downarrow$ & \textbf{RMSE} (floor) $\downarrow$ \\
        \hline
        STL & \textbf{81.19 ± 0.97} & 97.90 ± 0.24 & \textbf{0.2169 ± 0.0124} & 0.5428 ± 0.0206 \\
        MTL & 80.91 ± 0.57 & \textbf{97.94 ± 0.40} & 0.2180 ± 0.0093 & \textbf{0.5388 ± 0.0168} \\
    \hline
    \end{tabular}
    \caption{Comparison of single-task learning (STL) and multi-task learning (MTL). 
    The experiments are repeated four times, and the mean values of the metrics along with their standard deviations are reported. 
    The metrics used are Accuracy, Accuracy (±1), MAE, and  RMSE; $\uparrow$: The higher, the better; $\downarrow$: The lower, the better. 
    The best results are highlighted in \textbf{bold}.}
    \label{tab:stl_vs_mtl}
\end{table*}

\subsection{Single-task \textit{vs.} Multi-task Learning}\label{sec:comp_ex}

Adding an auxiliary task can, in many cases, enrich the feature representation learned for other related tasks. 
Visual inspection of the ground‐truth data shows that roofs are \emph{not} counted as floors when they are pitched or otherwise non-flat.  
We therefore introduced a complementary roof-type classification task: simplified to a binary label (flat / non-flat) derived from the Munich LoD2 database. 
We expect it to supply cues about façade structure. 

To test this hypothesis, we attached a single binary head to our network and trained it jointly with the main floor-count branch.  
Table~\ref{tab:stl_vs_mtl} compares single-task learning (STL) with this multi-task learning (MTL) variant.  
Overall differences are small: the MTL model matches STL in strict accuracy yet yields a marginal gain in Accuracy\,(±1) and a slightly lower MAE.  
The confusion matrices in Fig.~\ref{fig:confusion_matrix} provide further insight: MTL predictions for the 3–6-floor range cluster more tightly on the diagonal, suggesting that roof information helps disambiguate mid-rise façades.

Fig.~\ref{fig:grid_images} and Fig.~\ref{fig:grid_images_continue} give qualitative examples.  
Each image displays three numbers: the ground‐truth class (floors – 1), the STL prediction, and the MTL prediction.  
Some buildings are better handled by MTL, others by STL, and most show negligible differences.  
Hence, while roof-type supervision introduces useful context, its impact is nuanced rather than decisive; selecting richer or more targeted auxiliary tasks remains an open research direction.

\input{sections/figure_grid}

\subsection{Scalability of SLI} 

The accuracy of the proposed method is very high; however, one potential limitation is the scalability in acquiring large-scale, high-coverage, and high-quality SLIs. 

In our study, we used two sources: Mapillary SLIs and self-captured images. 

\begin{figure}[h!]
    \centering
    \includegraphics[width=\textwidth]{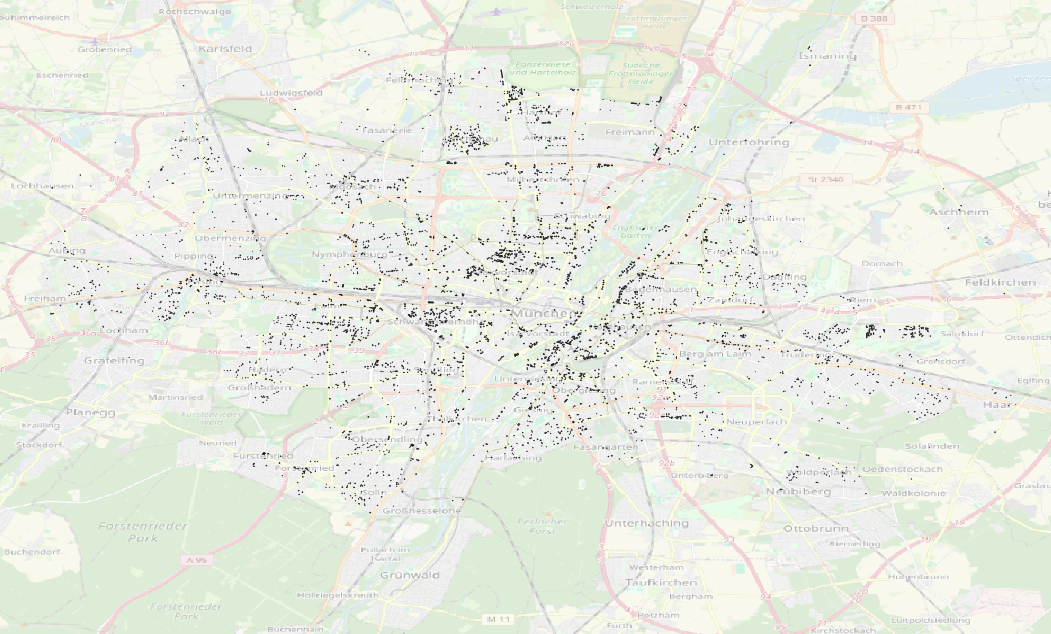}
    \caption{Geo-distribution of building footprints matched with final Mapillary subset images.}
    \label{fig:dataset_map}
\end{figure}

\emph{Mapillary} offers global coverage, convenient API access, and rich metadata (time, pose, camera model).  
However, many photos position the roadway centrally, leaving façades partly cropped or occluded, especially for high-rise buildings, where the building tops or bases were often cut off; after filtering, only 6 473 usable façade crops remained, representing 4 129 distinct buildings (Fig.~\ref{fig:dataset_map}).  
\emph{Self-captured} photos, in contrast, focused on underrepresented high-rise buildings and ensured that buildings were centered in the frame, reducing the need for detection and cropping. While this improved the diversity and generalizability of the dataset, the manual effort required limited the number of such images.

This limitation stems from the inherent nature of SLIs. To address this, one potential solution is to encourage volunteered geographic information (VGI) contributors to upload diverse and high-quality images, particularly of buildings not adjacent to roads or those occluded in typical street-level captures. Another promising approach is to combine SLIs with remote sensing data, such as high-resolution satellite imagery or aerial photos, to ensure complete building coverage and enhance model training. Integrating these data sources could mitigate occlusion and coverage issues, significantly improving scalability and accuracy.

%% file: sections/figure_grid.tex
\begin{figure*}[h!]
    \centering
    \setlength{\tabcolsep}{1pt}
    \renewcommand{\arraystretch}{0.8}
    \begin{tabular}{ccccccccc}
    \includegraphics[width=0.1\linewidth]{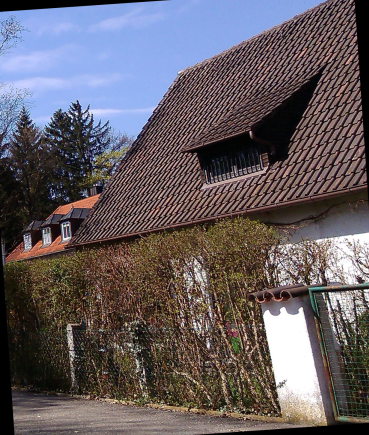}%

    & \includegraphics[width=0.1\linewidth]{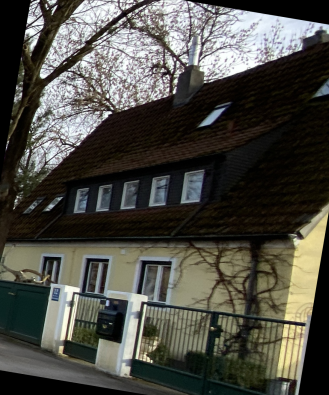}%

    & \includegraphics[width=0.1\linewidth]{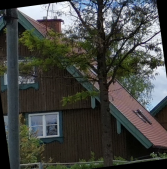}%

    & \includegraphics[width=0.1\linewidth]{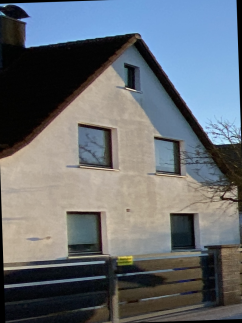}%

    & \includegraphics[width=0.1\linewidth]{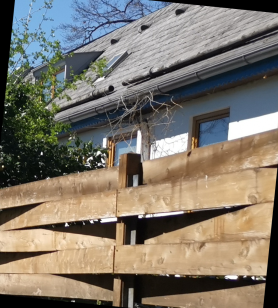}%

    & \includegraphics[width=0.1\linewidth]{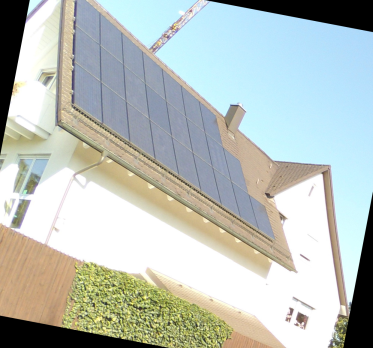}%

    & \includegraphics[width=0.1\linewidth]{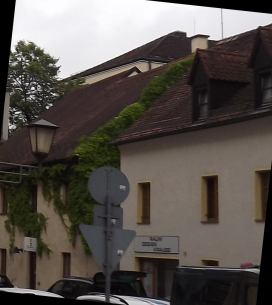}%

    & \includegraphics[width=0.1\linewidth]{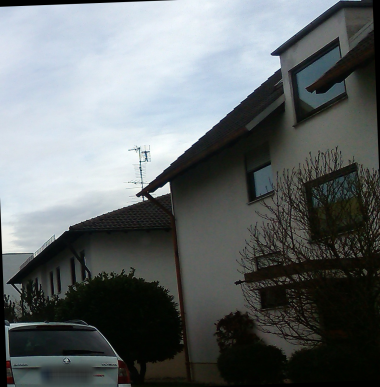}%

    & \includegraphics[width=0.1\linewidth]{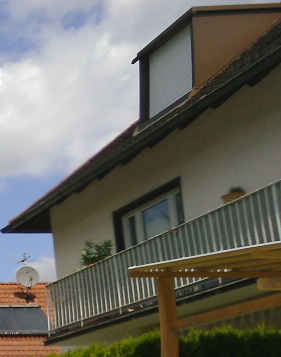}%
\\
    00,00,00&  00,00,01&  00,01,00&  00,01,01&  00,01,02&  00,02,01&  01,00,00&  01,00,01&  01,00,02\\
    \includegraphics[width=0.1\linewidth]{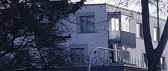}%

    & \includegraphics[width=0.1\linewidth]{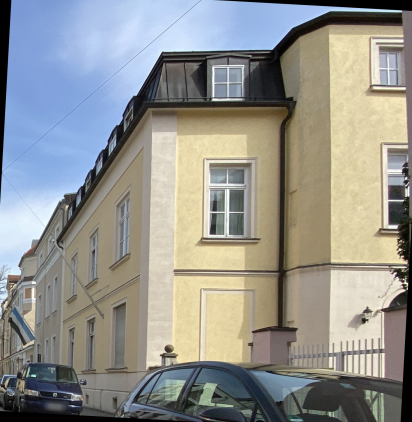}%

    & \includegraphics[width=0.1\linewidth]{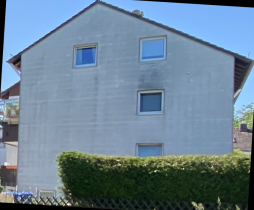}%

    & \includegraphics[width=0.1\linewidth]{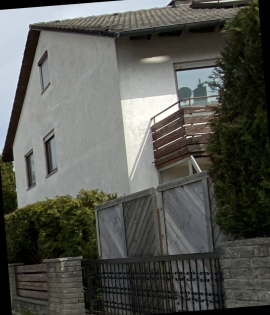}%

    & \includegraphics[width=0.1\linewidth]{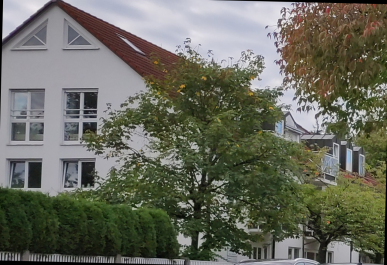}%

    & \includegraphics[width=0.1\linewidth]{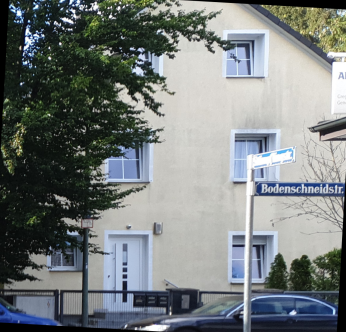}%

    & \includegraphics[width=0.1\linewidth,height=0.13\linewidth]{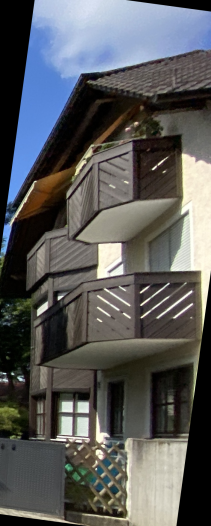}%

    & \includegraphics[width=0.1\linewidth,height=0.13\linewidth]{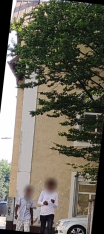}%

    & \includegraphics[width=0.1\linewidth]{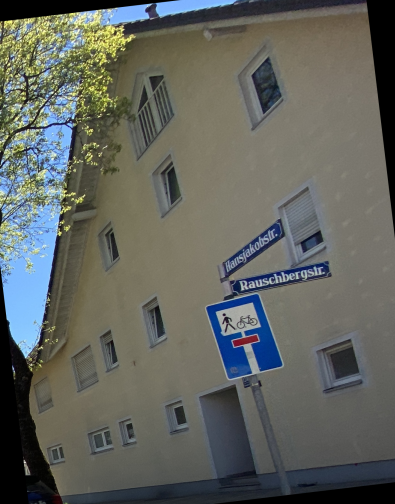}%
\\
    01,01,00&  01,01,01&  01,01,02&  01,02,00&  01,02,01&  01,02,02&  01,02,03&  01,03,02&  01,03,03\\
    \includegraphics[width=0.1\linewidth, height=0.13\linewidth]{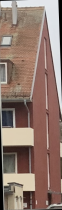}%

    & \includegraphics[width=0.1\linewidth, height=0.13\linewidth]{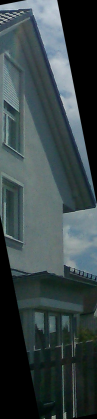}%

    & \includegraphics[width=0.1\linewidth]{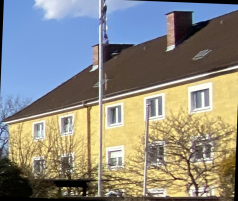}%

    & \includegraphics[width=0.1\linewidth]{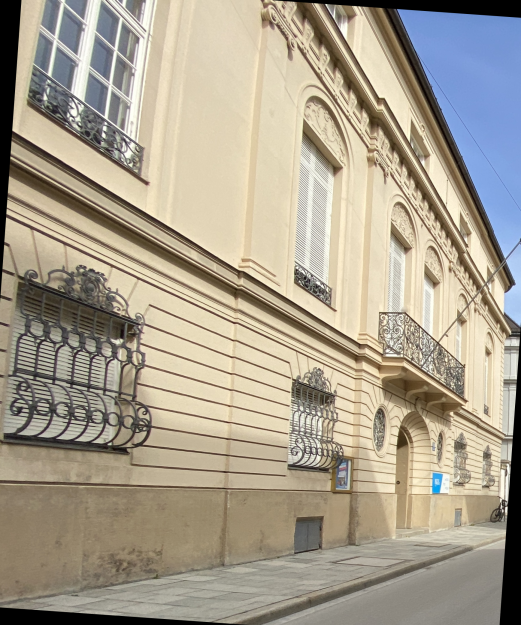}%

    & \includegraphics[width=0.1\linewidth]{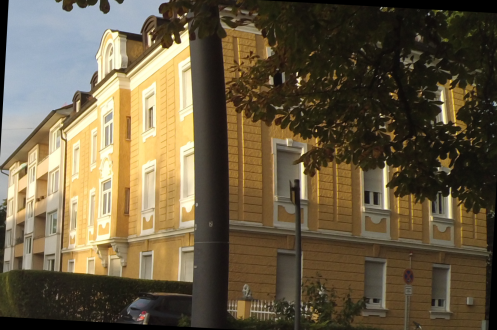}%

    & \includegraphics[width=0.1\linewidth]{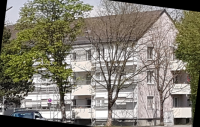}%

    & \includegraphics[width=0.1\linewidth]{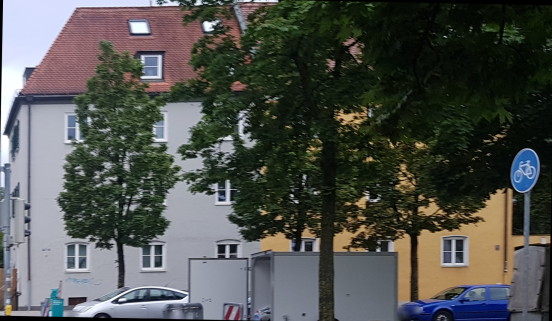}%

    & \includegraphics[width=0.1\linewidth]{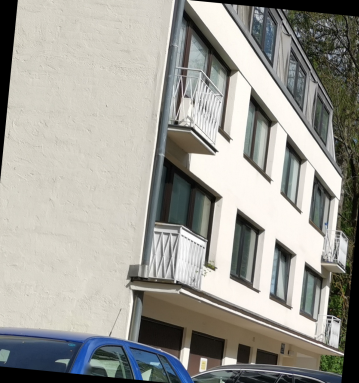}%

    & \includegraphics[width=0.1\linewidth]{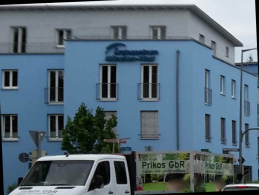}%
\\
    02,00,02&  02,01,00&  02,01,01&  02,01,02&  02,02,01&  02,02,02&  02,02,03&  02,03,02&  02,03,03\\
    \includegraphics[width=0.1\linewidth,height=0.13\linewidth]{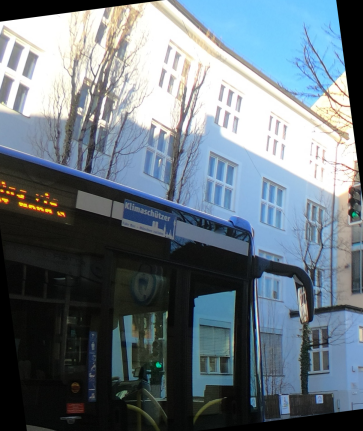}%

    & \includegraphics[width=0.1\linewidth]{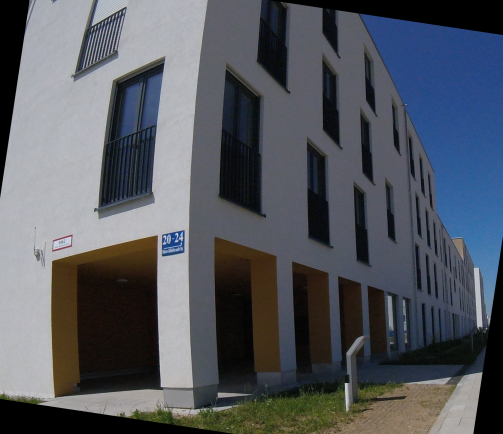}%

    & \includegraphics[width=0.1\linewidth]{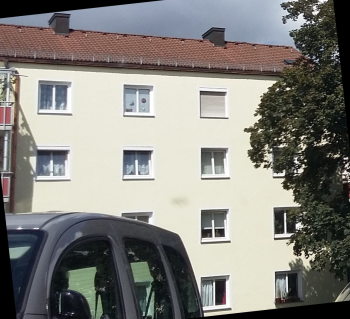}%

    & \includegraphics[width=0.1\linewidth]{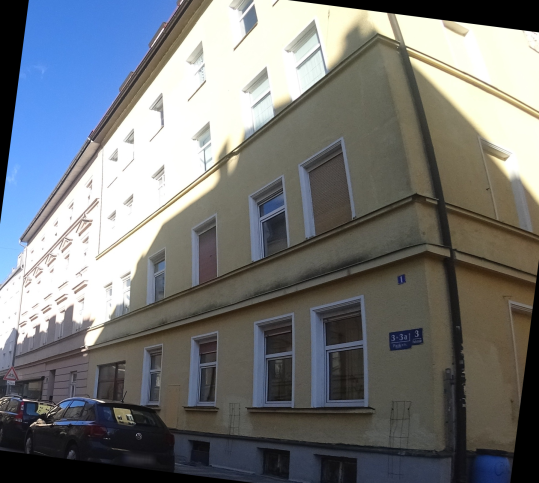}%

    & \includegraphics[width=0.1\linewidth]{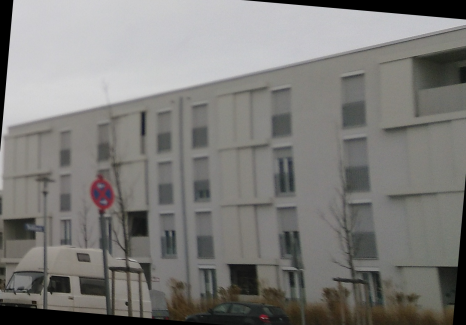}%

    & \includegraphics[width=0.1\linewidth]{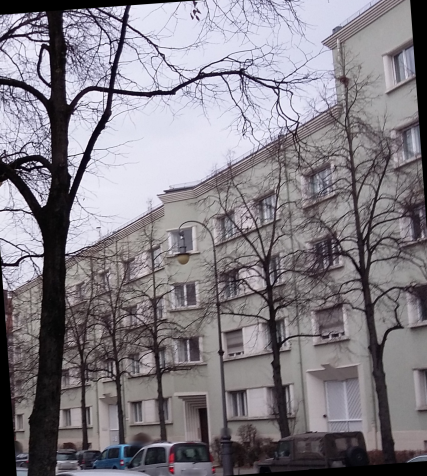}%

    & \includegraphics[width=0.1\linewidth]{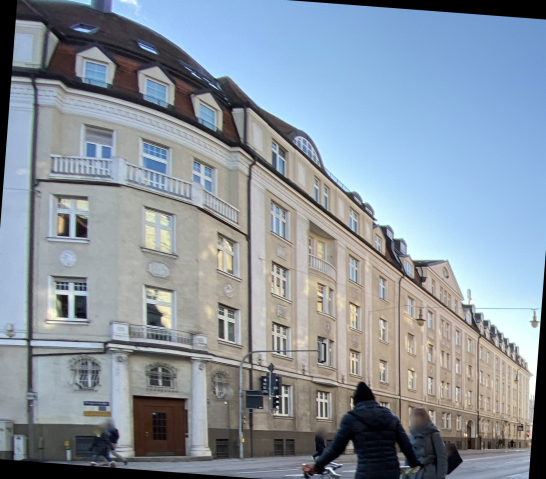}%

    & \includegraphics[width=0.1\linewidth, height=0.13\linewidth]{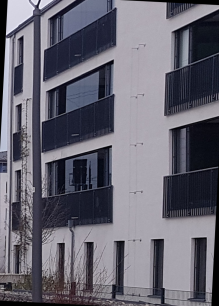}%

    & \includegraphics[width=0.1\linewidth]{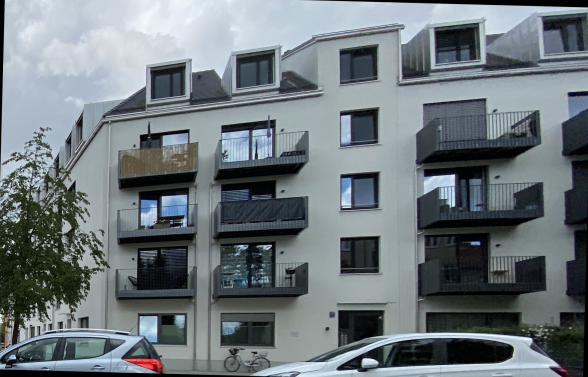}%
\\
    03,01,03&  03,02,02&  03,02,03&  03,03,02&  03,03,03&  03,03,04&  03,03,05&  03,04,02&  03,04,03\\
    \includegraphics[width=0.1\linewidth,height=0.13\linewidth]{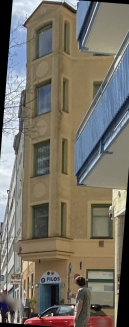}%

    & \includegraphics[width=0.1\linewidth,height=0.13\linewidth]{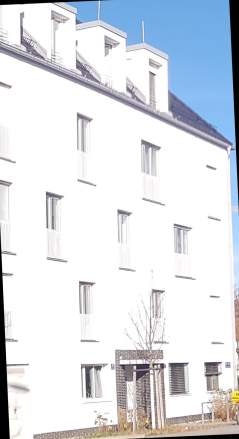}%

    & \includegraphics[width=0.1\linewidth]{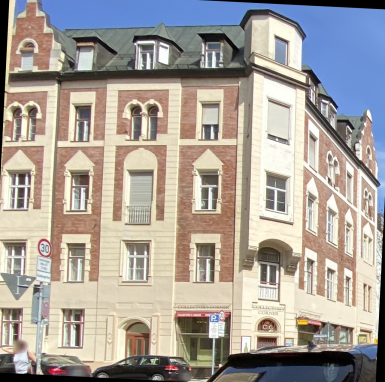}%

    & \includegraphics[width=0.1\linewidth,height=0.13\linewidth]{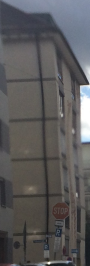}%

    & \includegraphics[width=0.1\linewidth,height=0.13\linewidth]{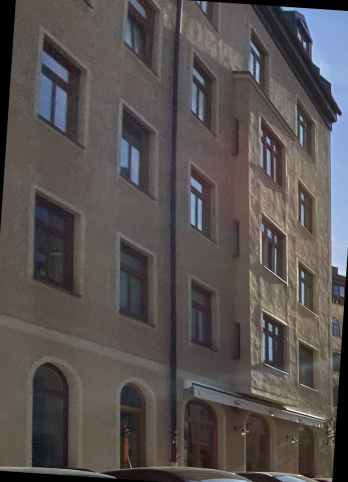}%

    & \includegraphics[width=0.1\linewidth,height=0.13\linewidth]{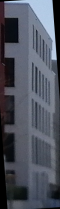}%

    & \includegraphics[width=0.1\linewidth]{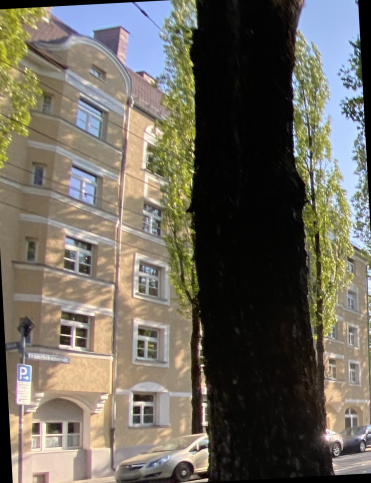}%

    & \includegraphics[width=0.1\linewidth]{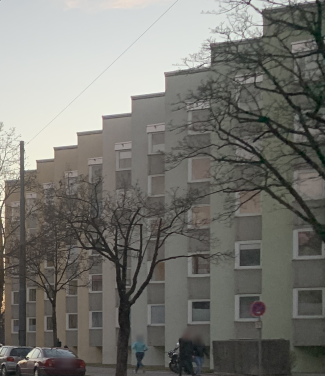}%

    & \includegraphics[width=0.1\linewidth]{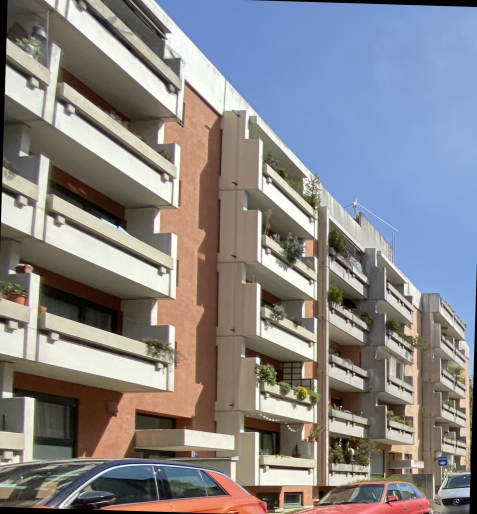}%
\\
    03,04,04&  04,02,04&  04,03,03&  04,03,04&  04,04,03&  04,04,04&  04,04,05&  04,05,03&  04,05,04\\
    \includegraphics[width=0.1\linewidth,height=0.13\linewidth]{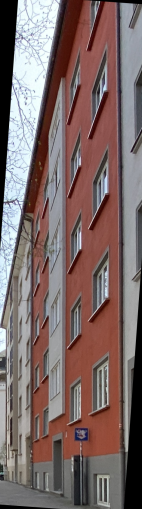}%

    & \includegraphics[width=0.1\linewidth,height=0.13\linewidth]{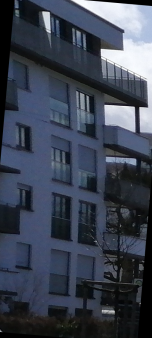}%

    & \includegraphics[width=0.1\linewidth]{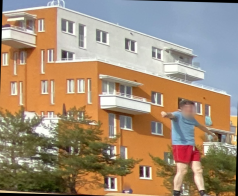}%

    & \includegraphics[width=0.1\linewidth]{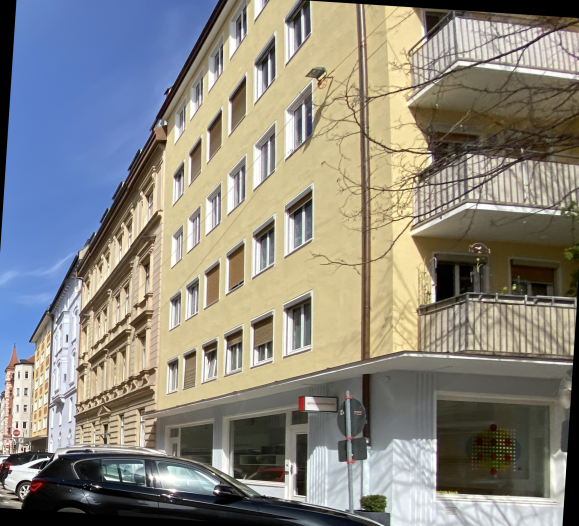}%

    & \includegraphics[width=0.1\linewidth,height=0.13\linewidth]{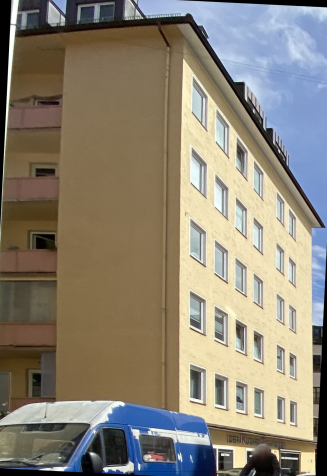}%

    & \includegraphics[width=0.1\linewidth]{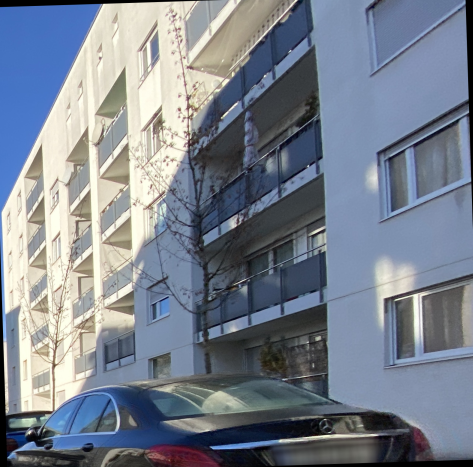}%

    & \includegraphics[width=0.1\linewidth]{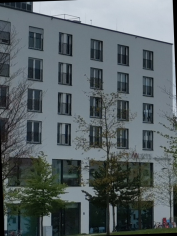}%

    & \includegraphics[width=0.1\linewidth,height=0.13\linewidth]{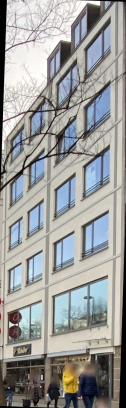}%

    & \includegraphics[width=0.1\linewidth]{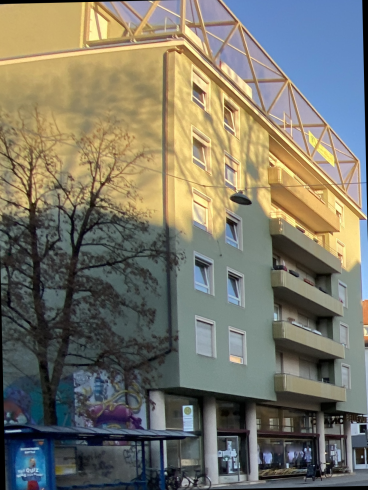}%
\\
    04,05,05&  05,03,03&  05,03,04&  05,04,04&  05,04,05&  05,05,04&  05,05,05&  05,05,06&  05,05,07\\
    \includegraphics[width=0.1\linewidth]{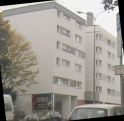}%

    & \includegraphics[width=0.1\linewidth]{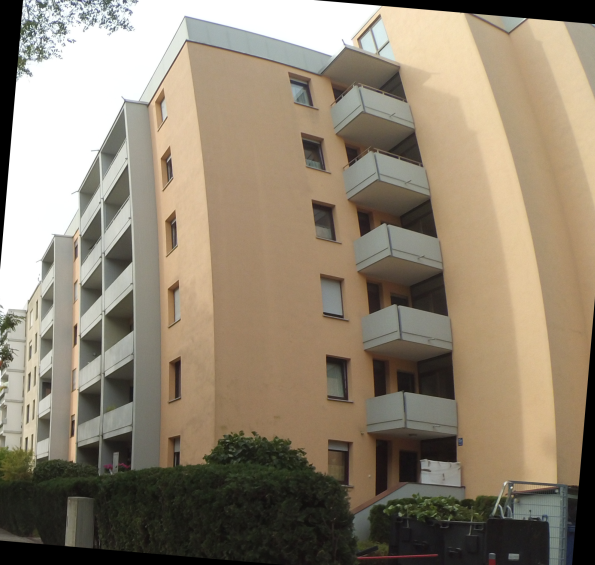}%

    & \includegraphics[width=0.1\linewidth]{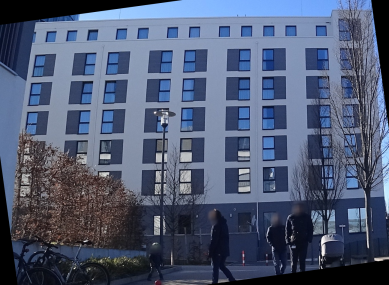}%

    & \includegraphics[width=0.1\linewidth]{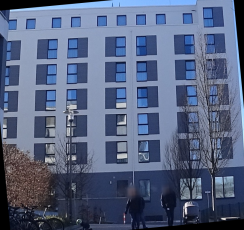}%

    & \includegraphics[width=0.1\linewidth]{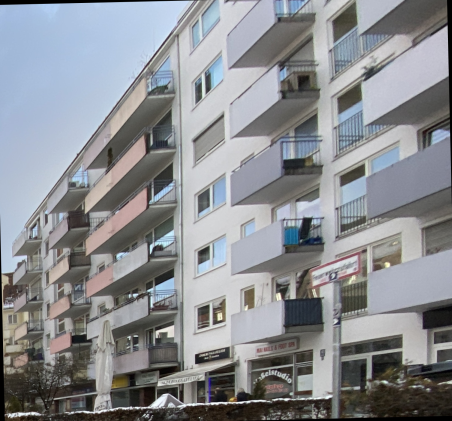}%

    & \includegraphics[width=0.1\linewidth]{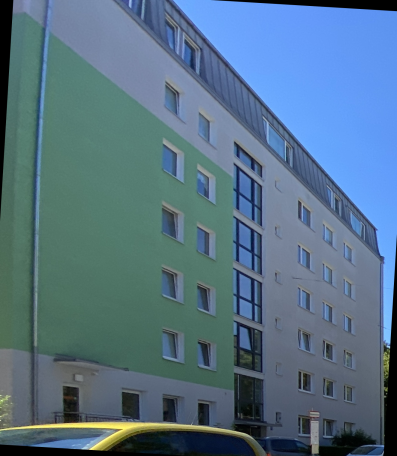}%

    & \includegraphics[width=0.1\linewidth]{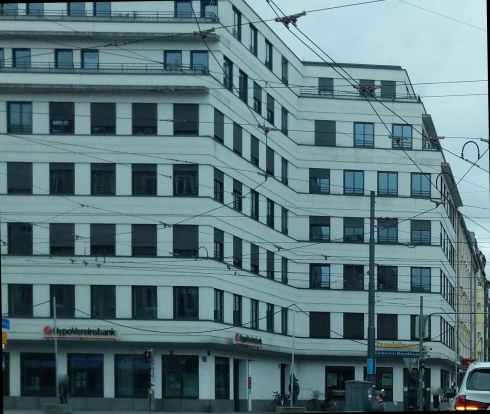}%

    & \includegraphics[width=0.1\linewidth,height=0.13\linewidth]{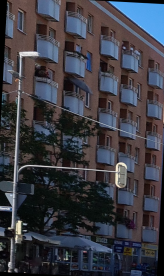}%

    & \includegraphics[width=0.1\linewidth]{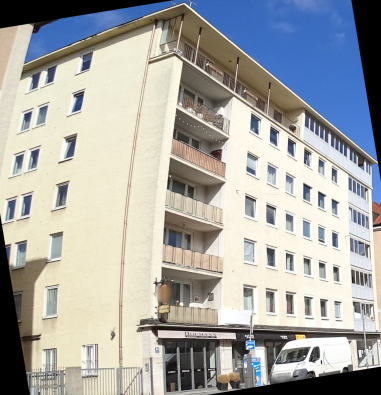}%
\\
    05,06,05&  05,06,06&  05,06,07&  05,07,07&  06,05,05&  06,05,06&  06,06,05&  06,06,06&  06,07,08\\
\end{tabular}
    \caption{Examples of buildings with ground truth, STL, and MTL predictions. The three numbers under each building image indicate the ground truth floor class (floor number - 1), the STL result, and the MTL result, respectively.}
    \label{fig:grid_images}
\end{figure*}

\begin{figure*}
\centering
    \setlength{\tabcolsep}{1pt}
    \renewcommand{\arraystretch}{0.8}
    \begin{tabular}{ccccccccc}
    \includegraphics[width=0.1\linewidth]{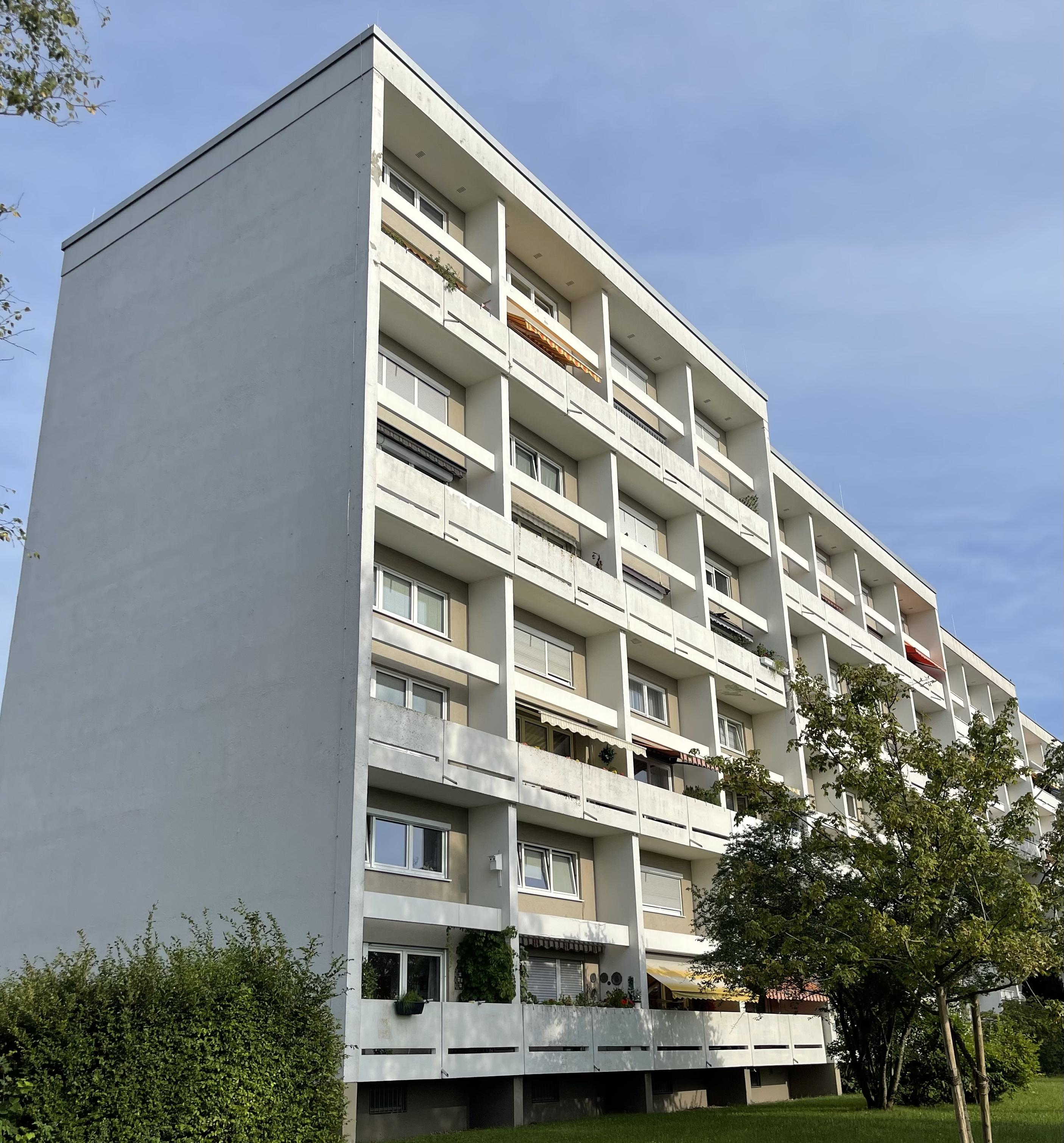}%

    & \includegraphics[width=0.1\linewidth,height=0.13\linewidth]{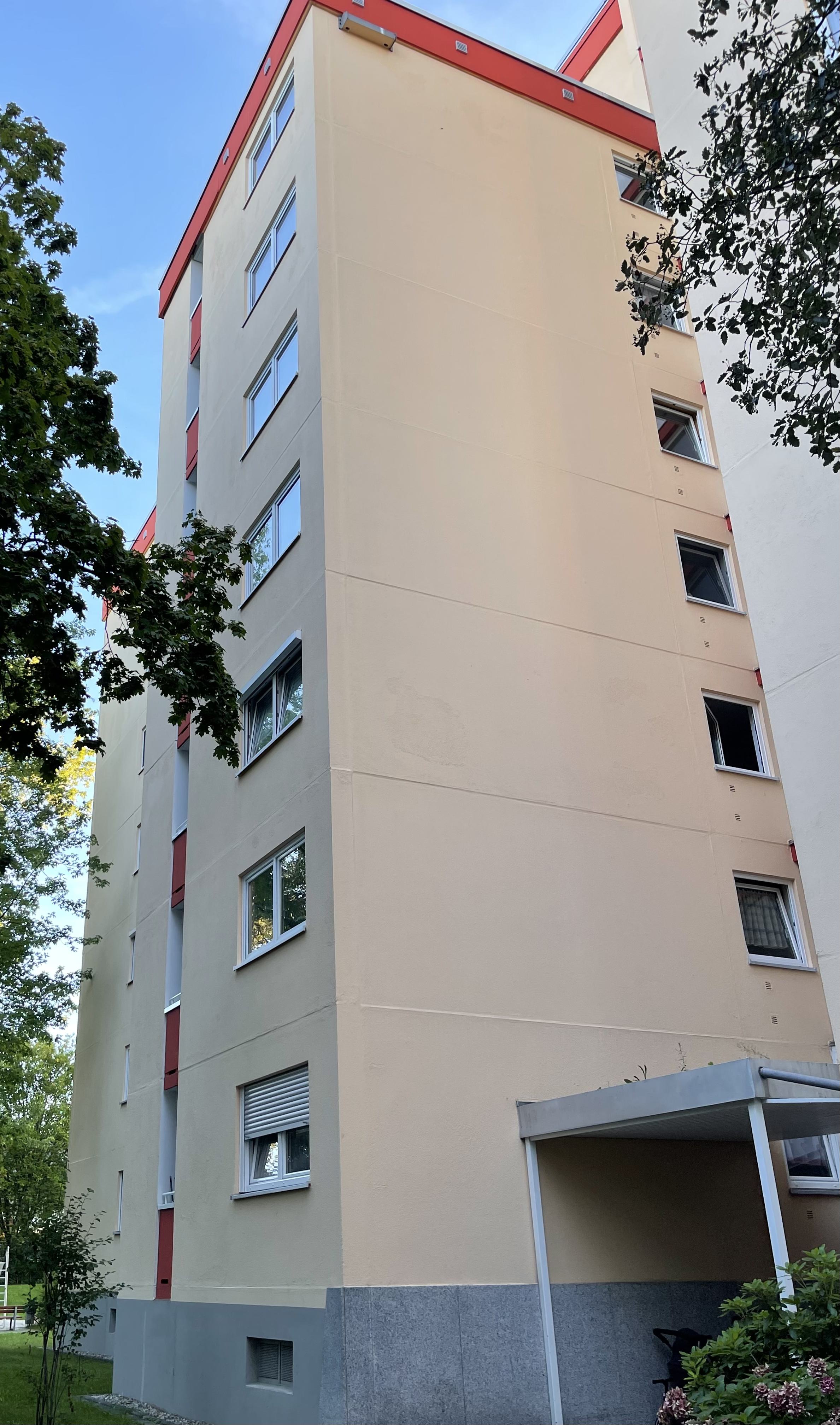}%

    & \includegraphics[width=0.1\linewidth]{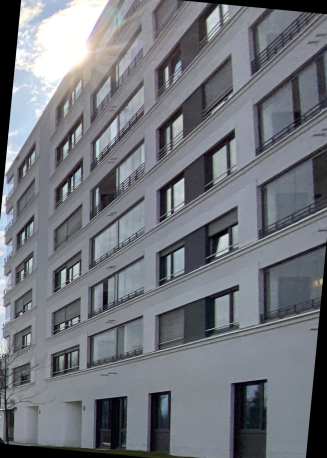}%

    & \includegraphics[width=0.1\linewidth]{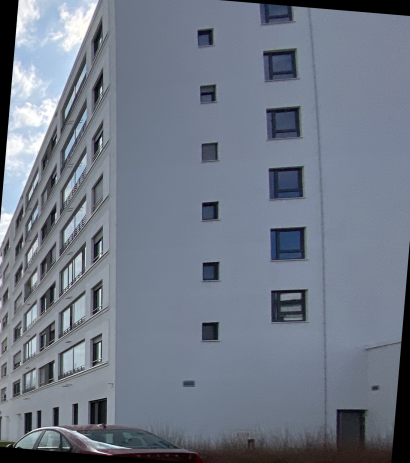}%

    & \includegraphics[width=0.1\linewidth]{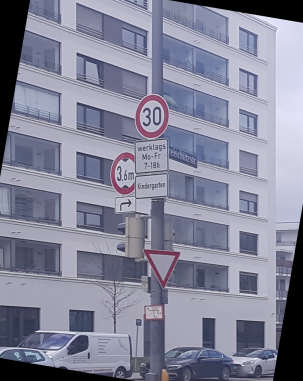}%

    & \includegraphics[width=0.1\linewidth]{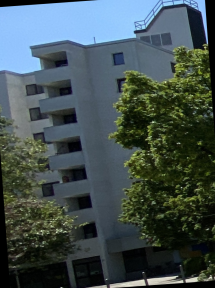}%

    & \includegraphics[width=0.1\linewidth]{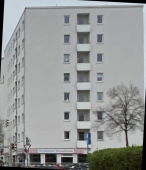}%

    & \includegraphics[width=0.1\linewidth, height=0.13\linewidth]{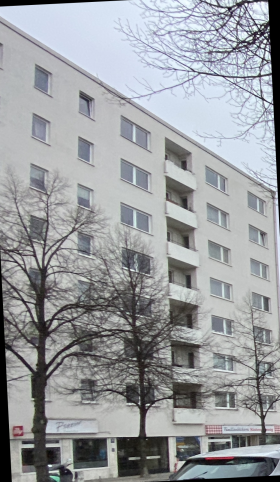}%

    & \includegraphics[width=0.1\linewidth, height=0.13\linewidth]{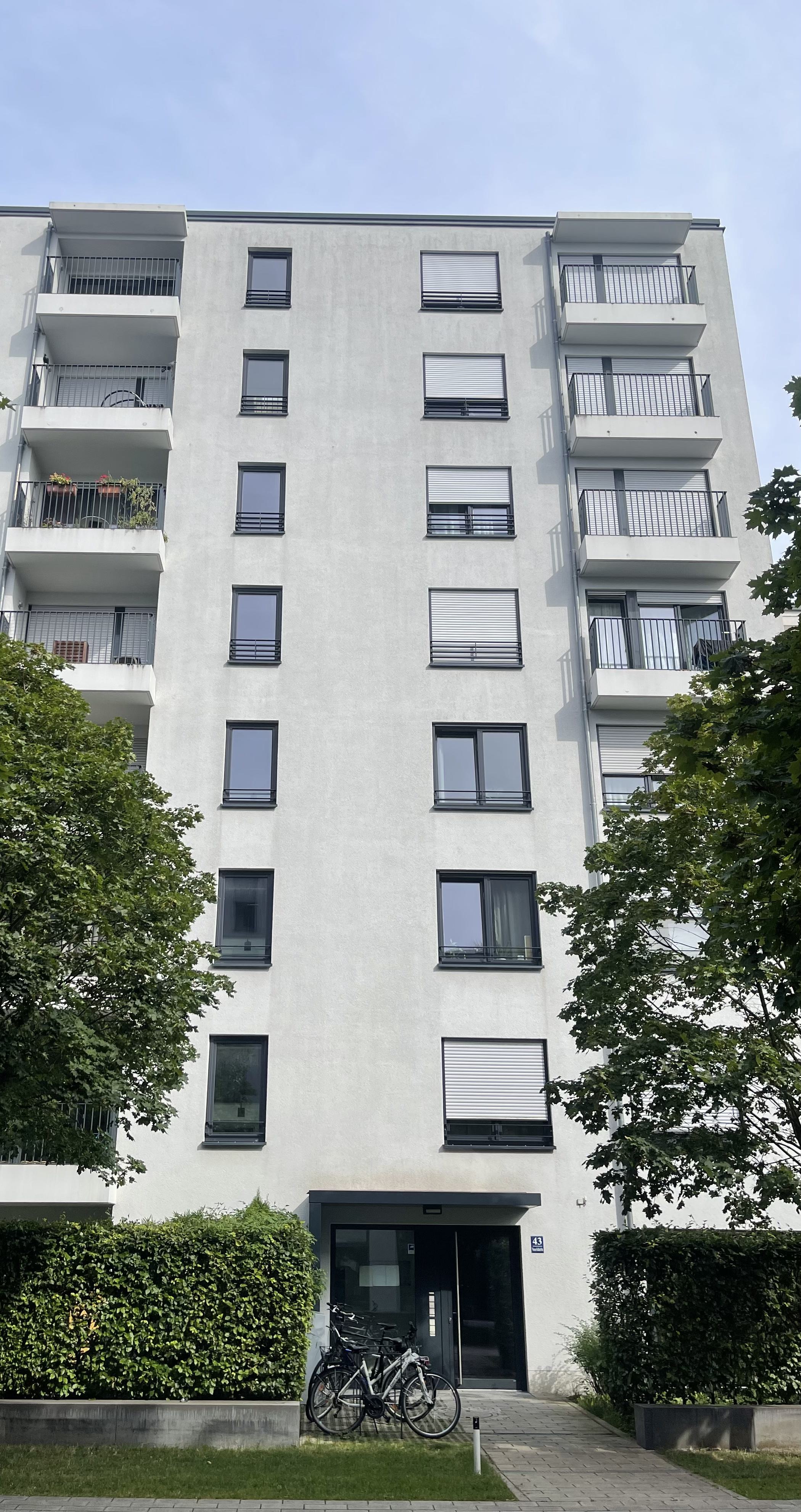}%
\\
    07,03,03&  07,04,08&  07,05,05&  07,06,05&  07,06,06&  07,06,07&  07,07,07&  07,07,08&  07,08,08\\
    \includegraphics[width=0.1\linewidth]{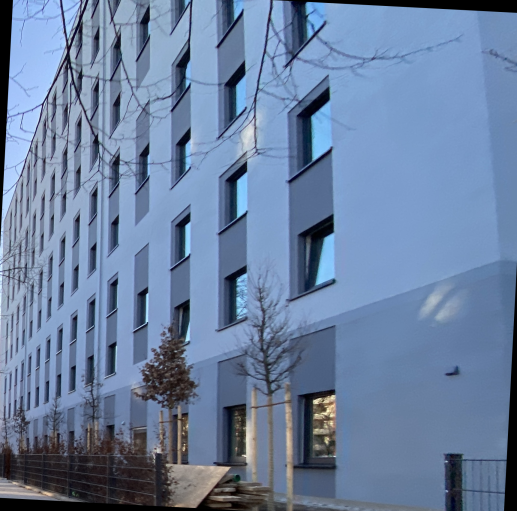}%

    & \includegraphics[width=0.1\linewidth]{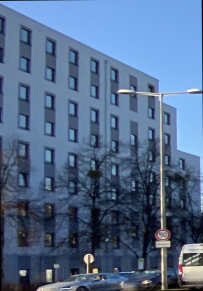}%

    & \includegraphics[width=0.1\linewidth]{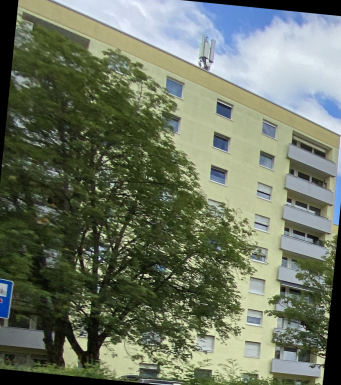}%

    & \includegraphics[width=0.1\linewidth]{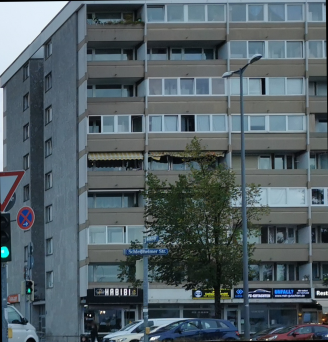}%

    & \includegraphics[width=0.1\linewidth]{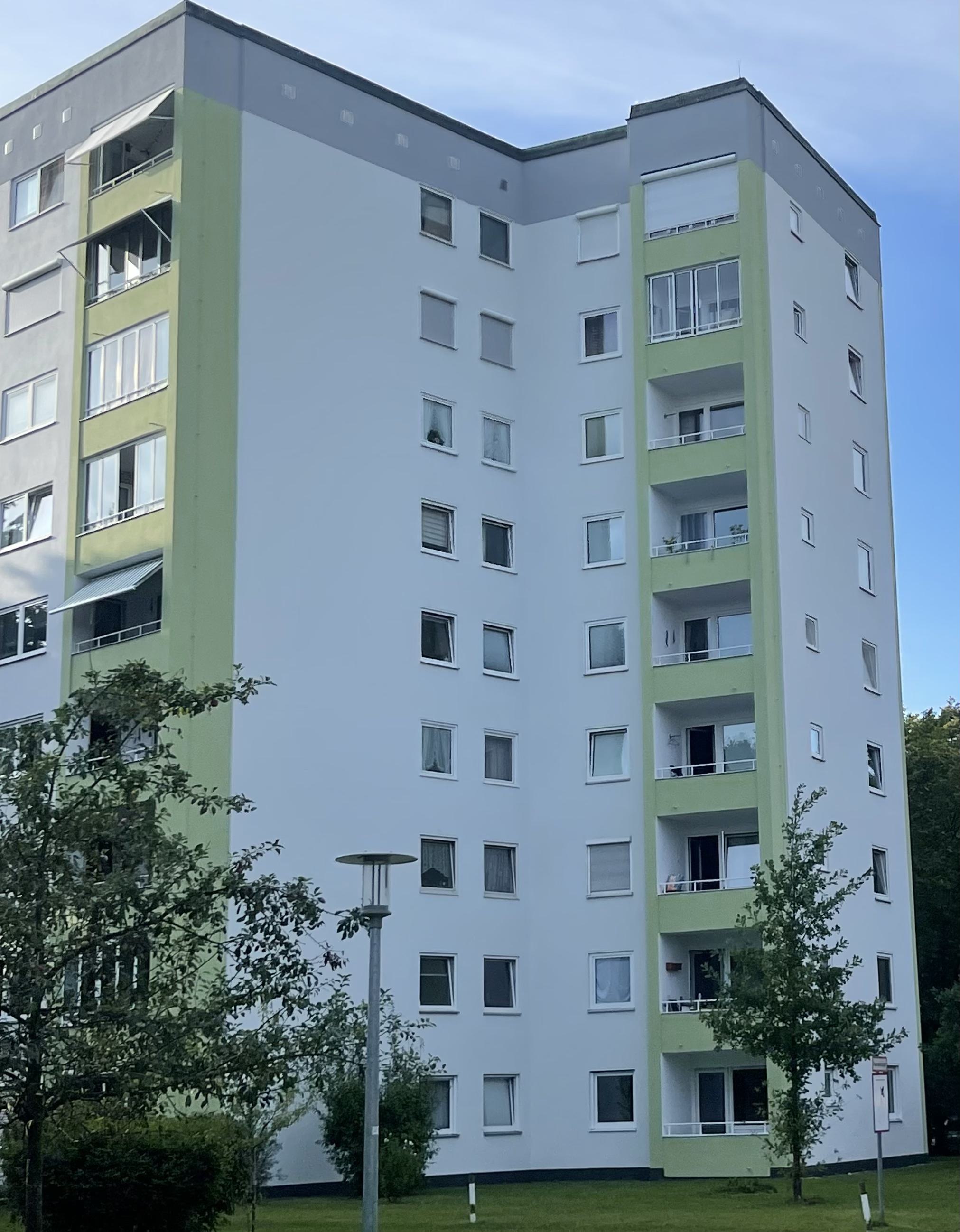}%

    & \includegraphics[width=0.1\linewidth]{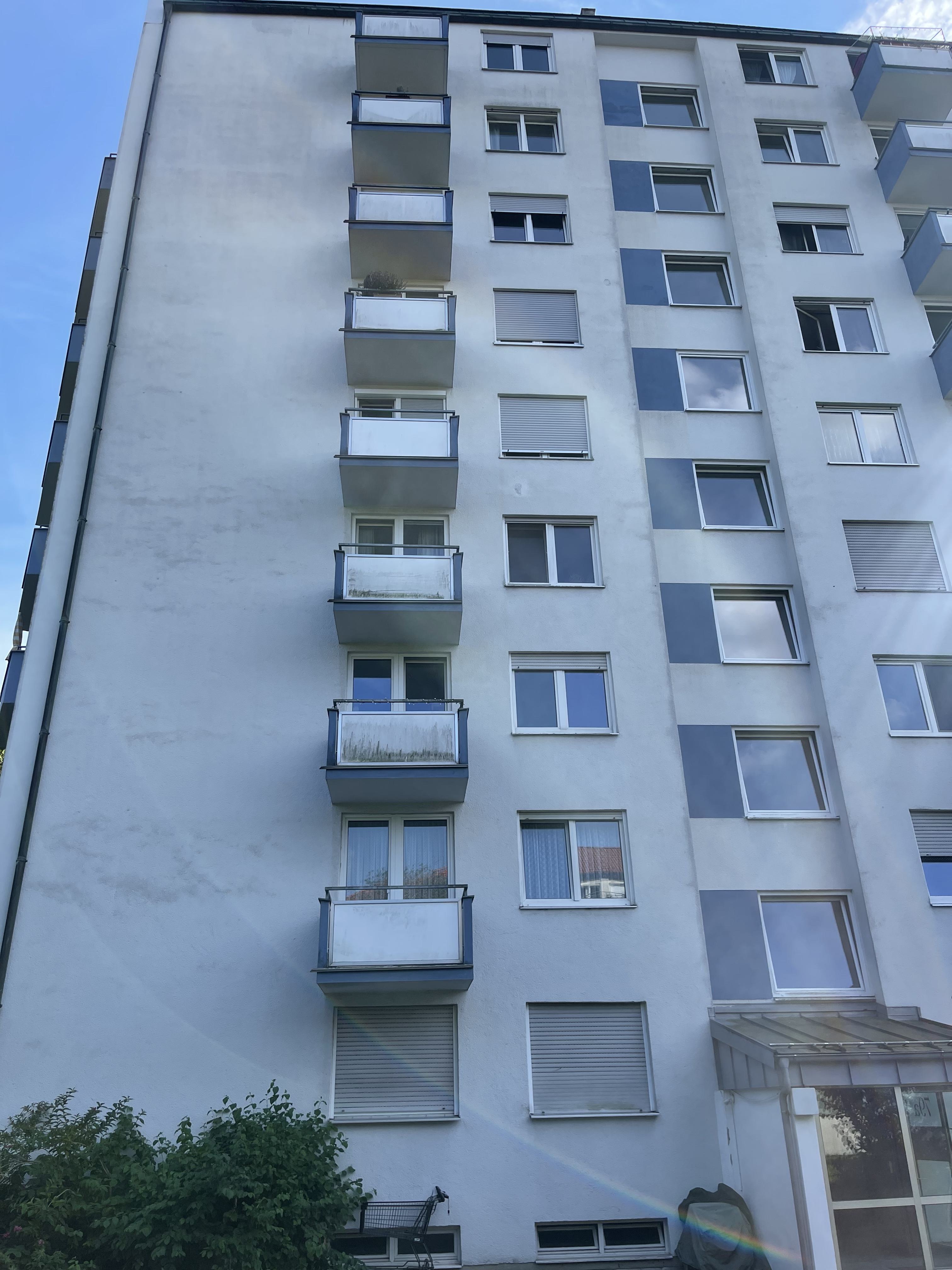}%

    & \includegraphics[width=0.1\linewidth]{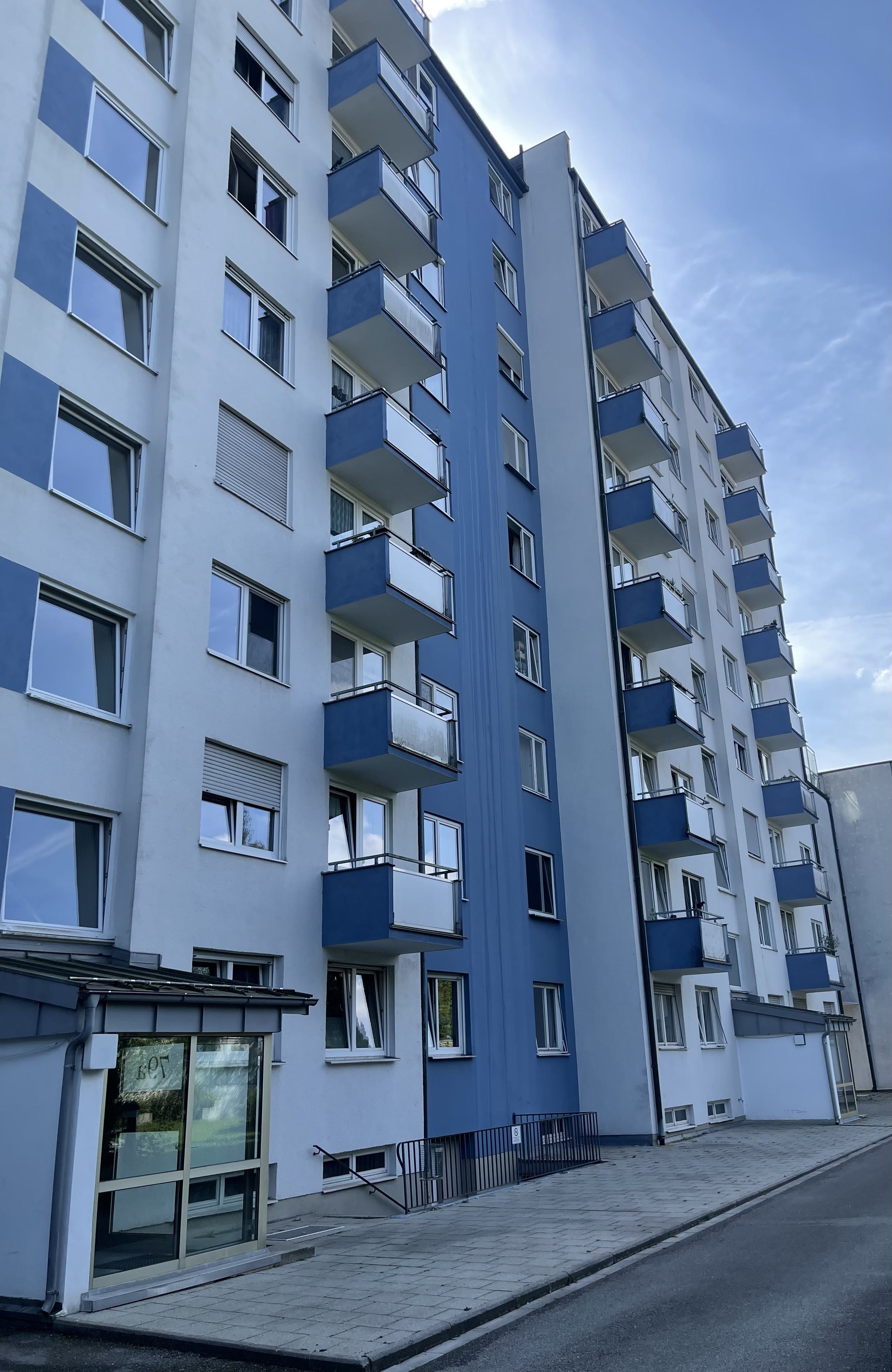}%

    & \includegraphics[width=0.1\linewidth]{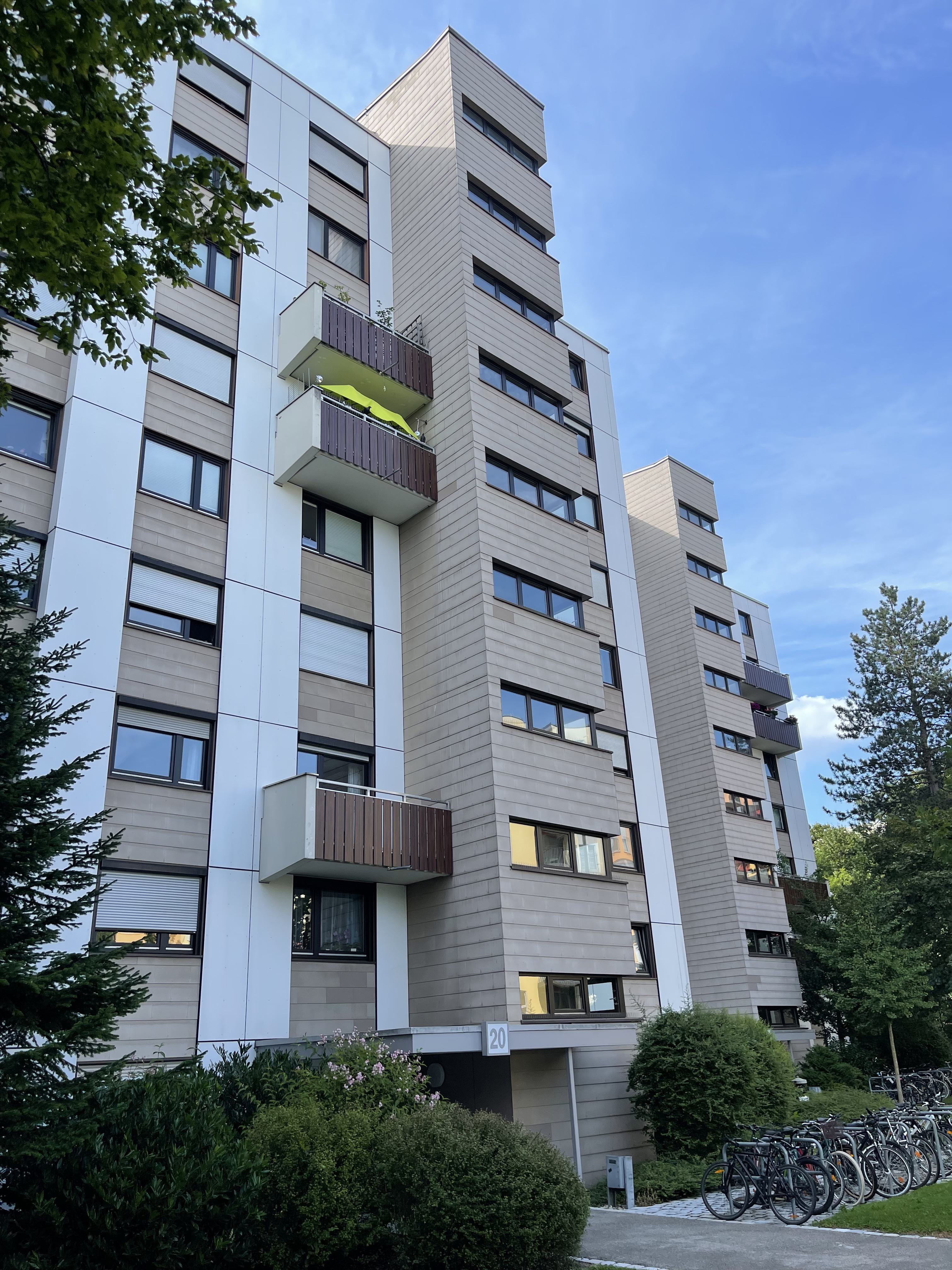}%

    & \includegraphics[width=0.1\linewidth]{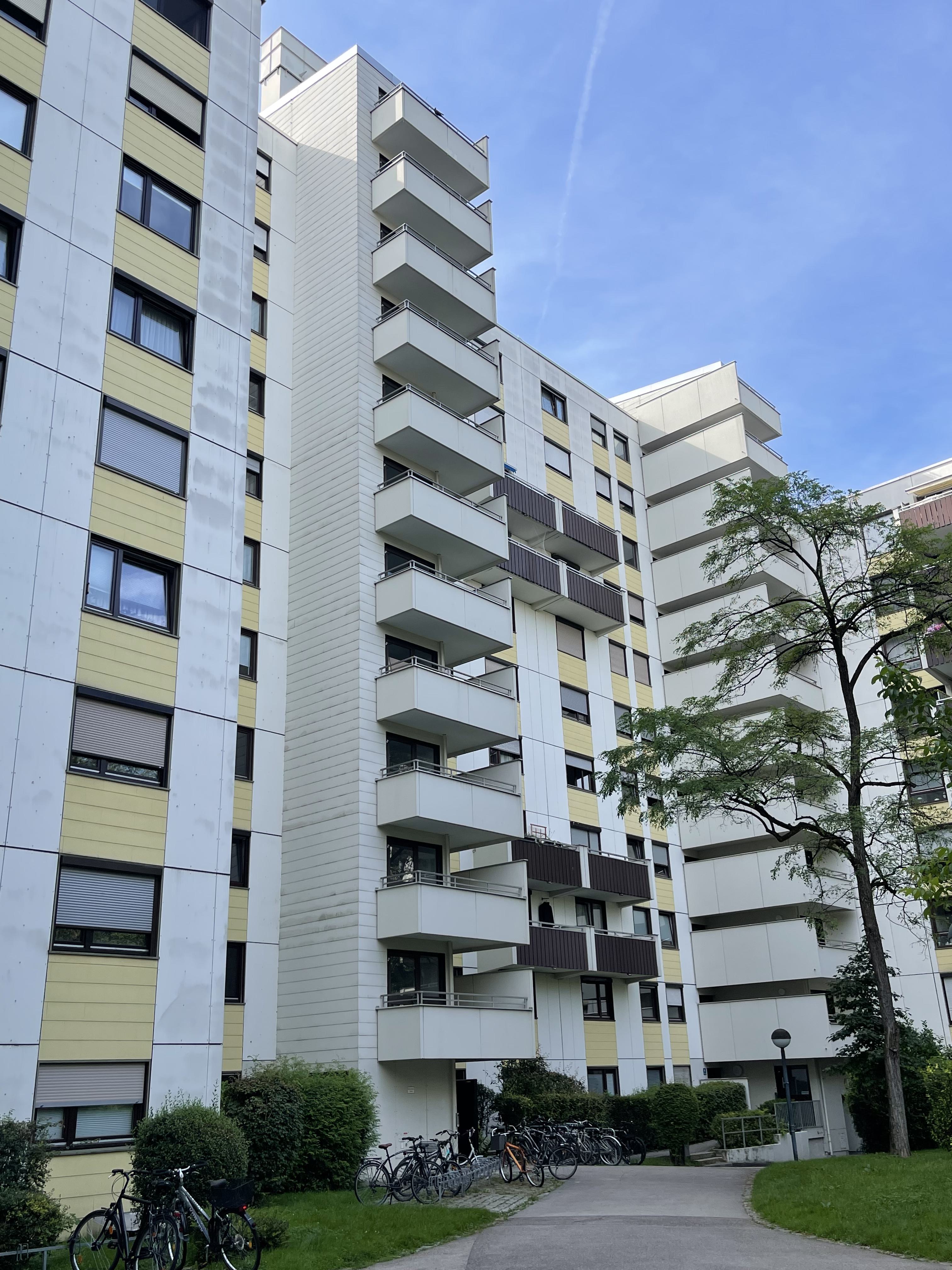}%
\\
    08,04,04&  08,05,07&  08,07,07&  08,07,08&  08,08,08&  09,08,08&  09,09,08&  09,09,09&  09,10,10\\
    \includegraphics[width=0.1\linewidth]{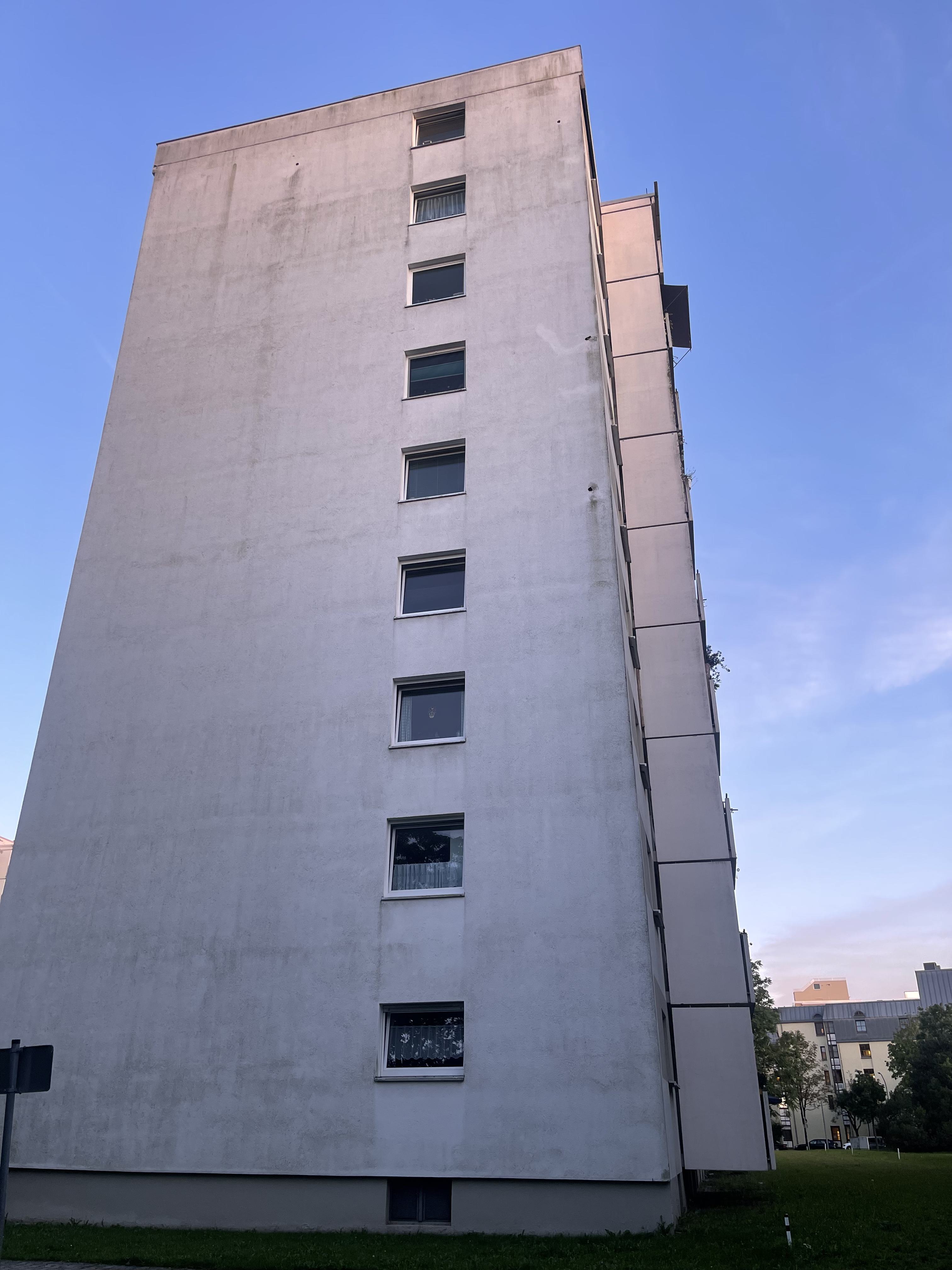}%

    & \includegraphics[width=0.1\linewidth]{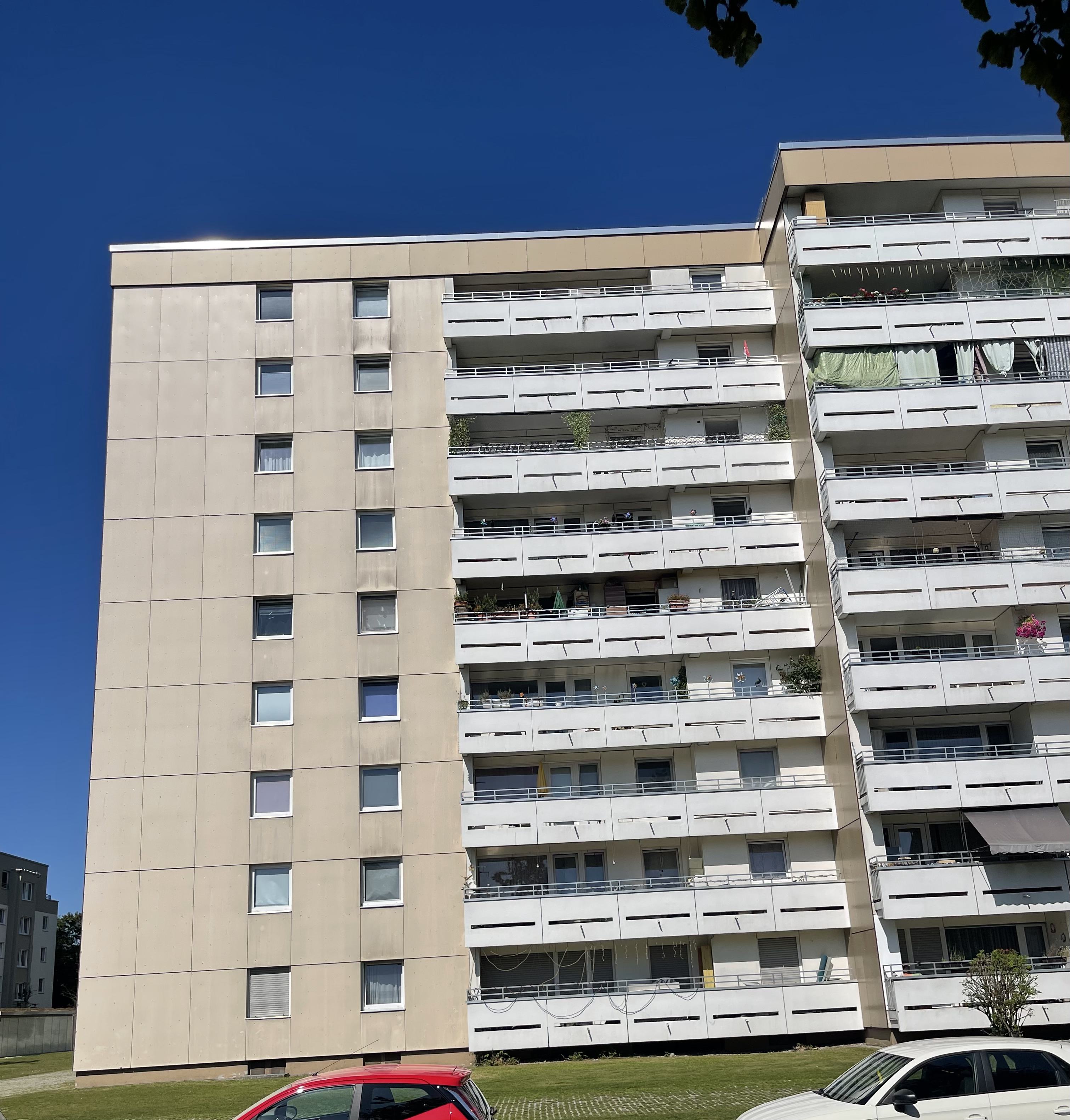}%

    & \includegraphics[width=0.1\linewidth]{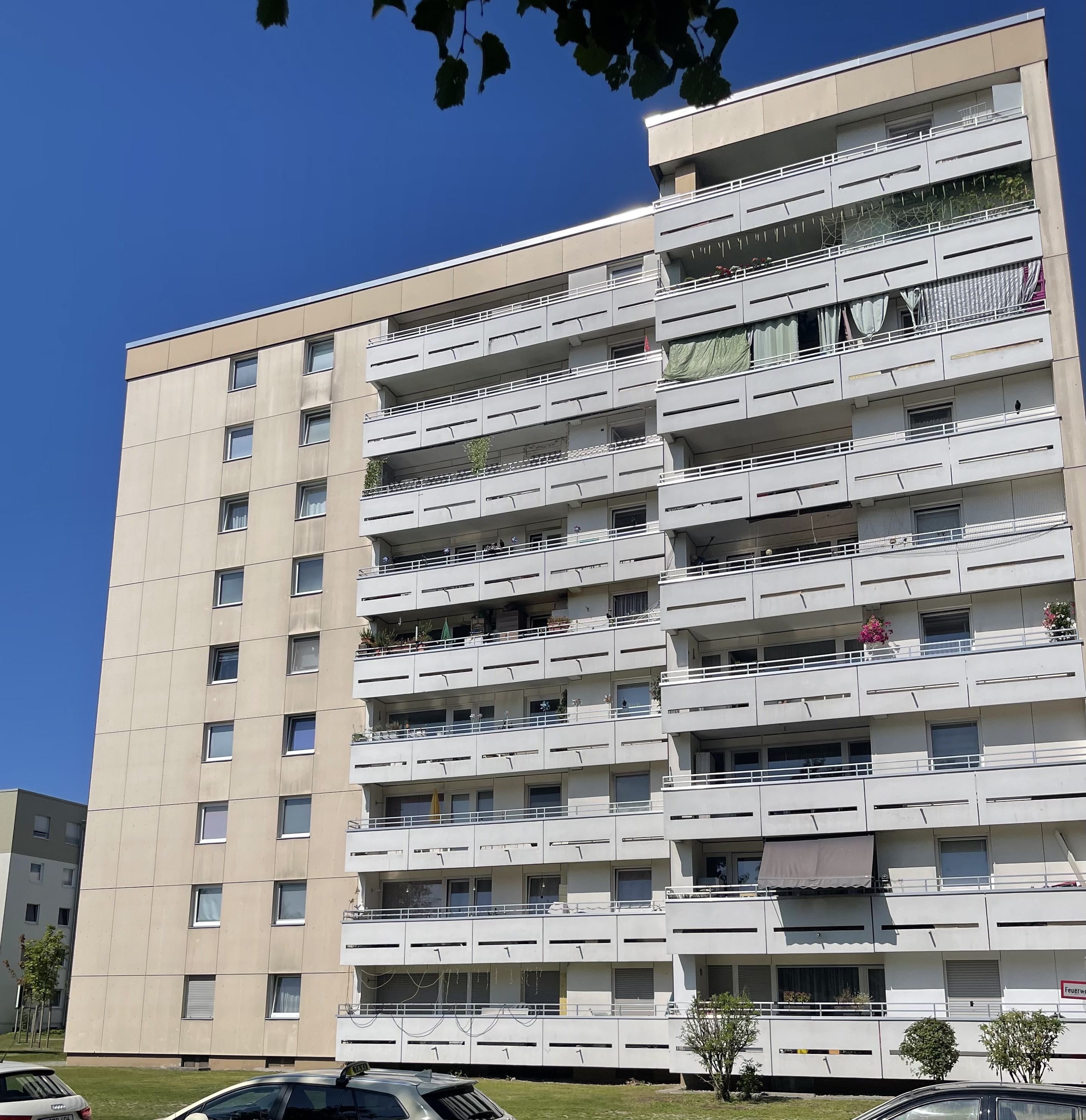}%

    & \includegraphics[width=0.1\linewidth]{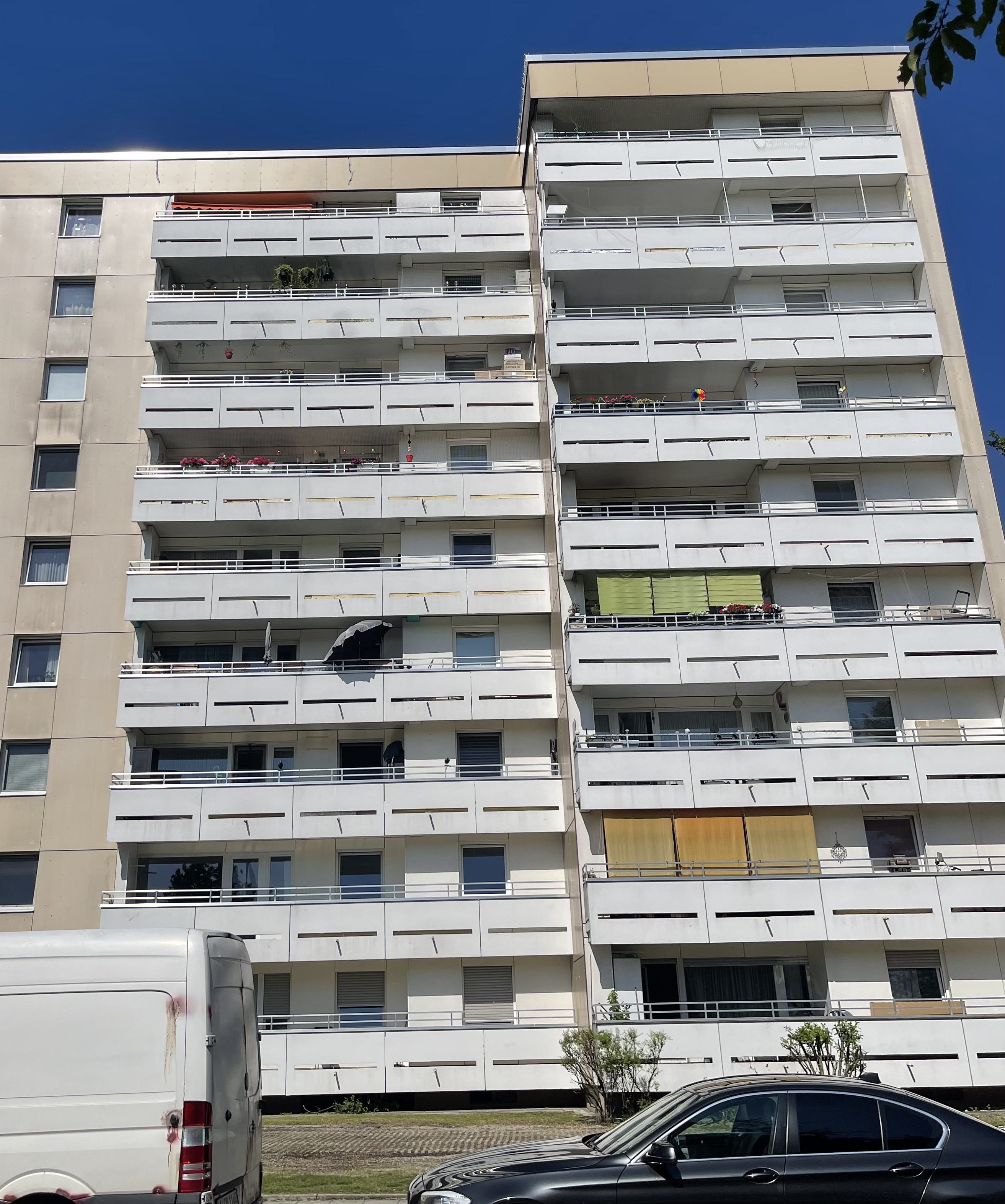}%

    & \includegraphics[width=0.1\linewidth]{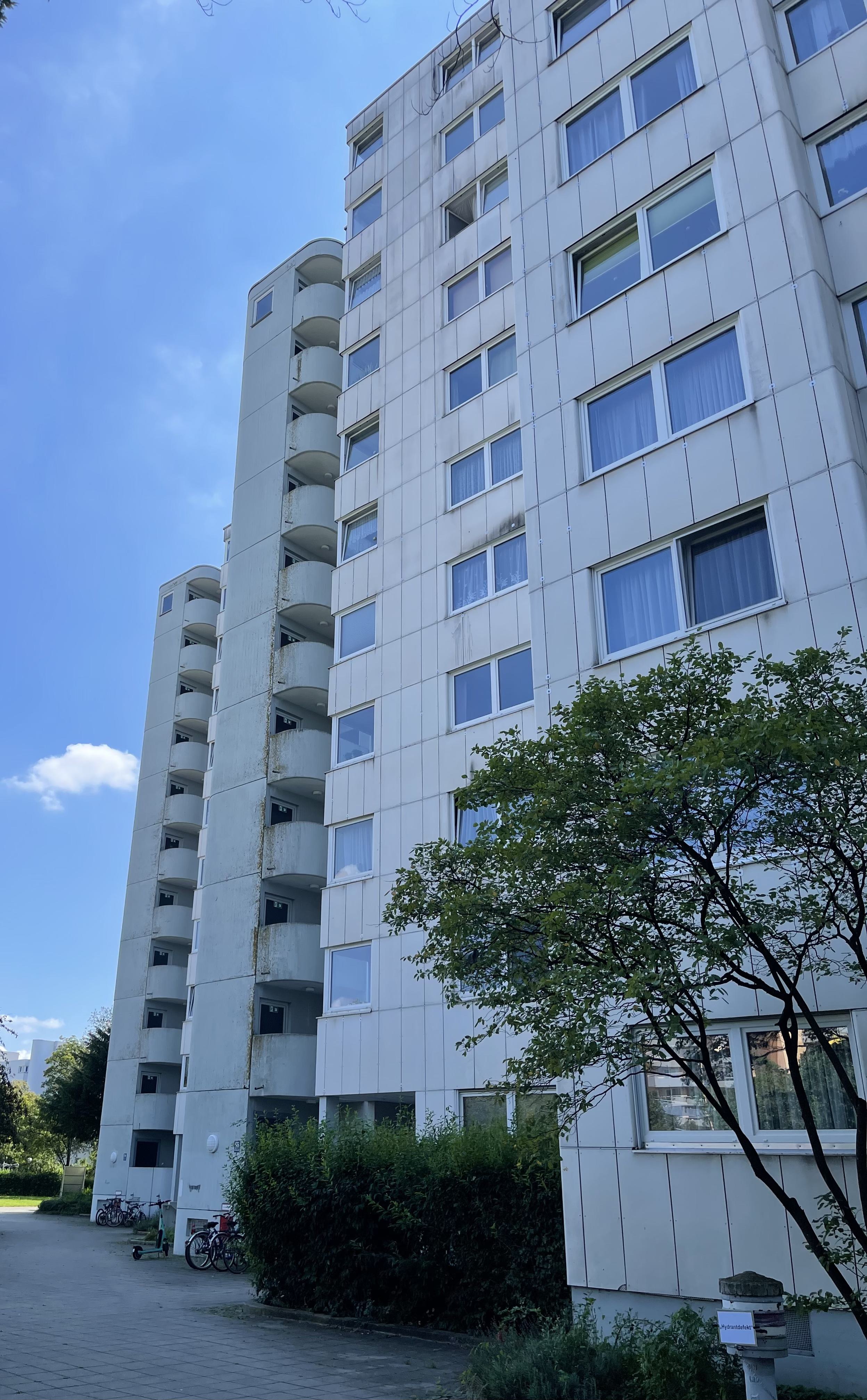}%

    & \includegraphics[width=0.1\linewidth]{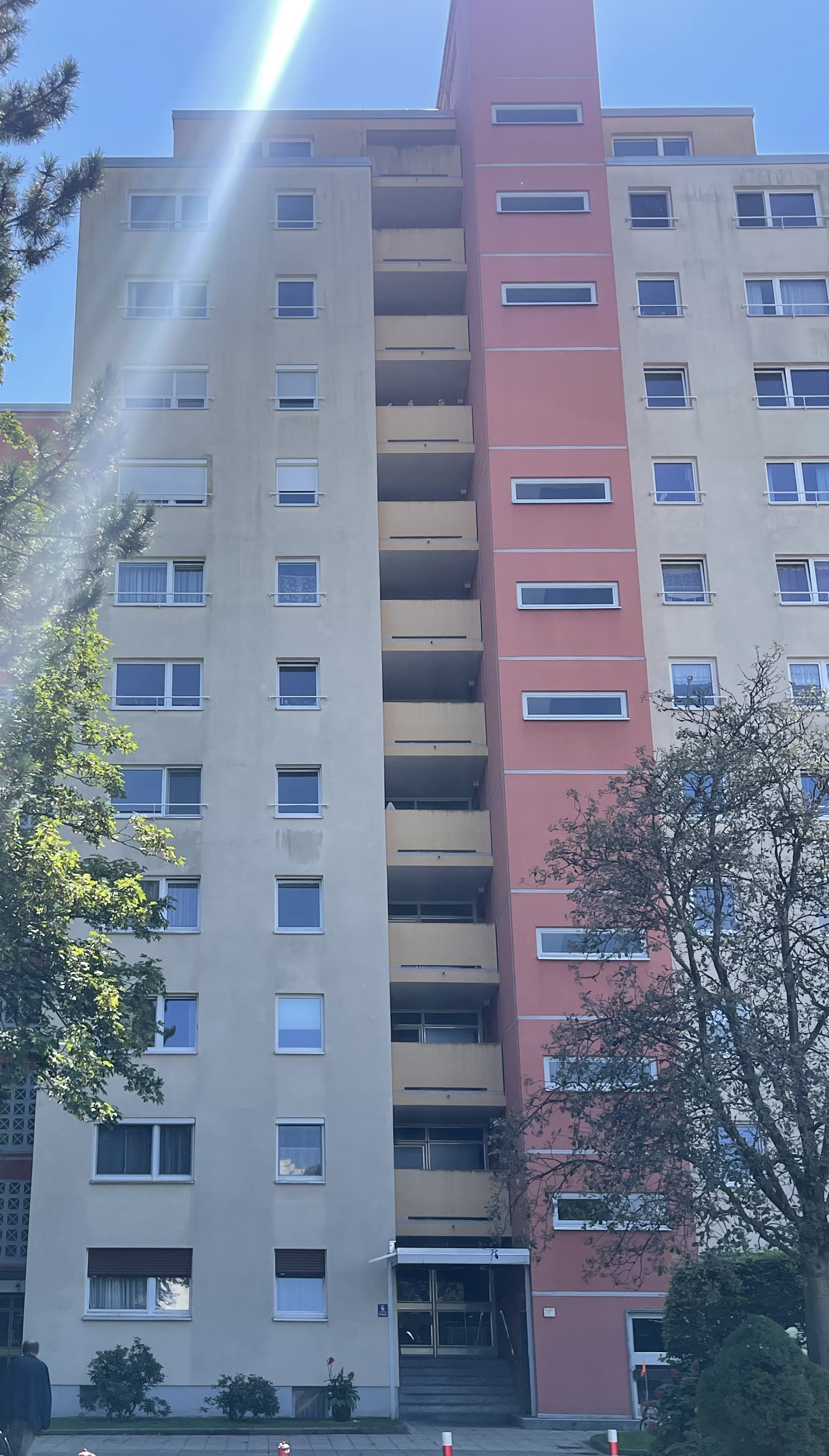}%

    & \includegraphics[width=0.1\linewidth]{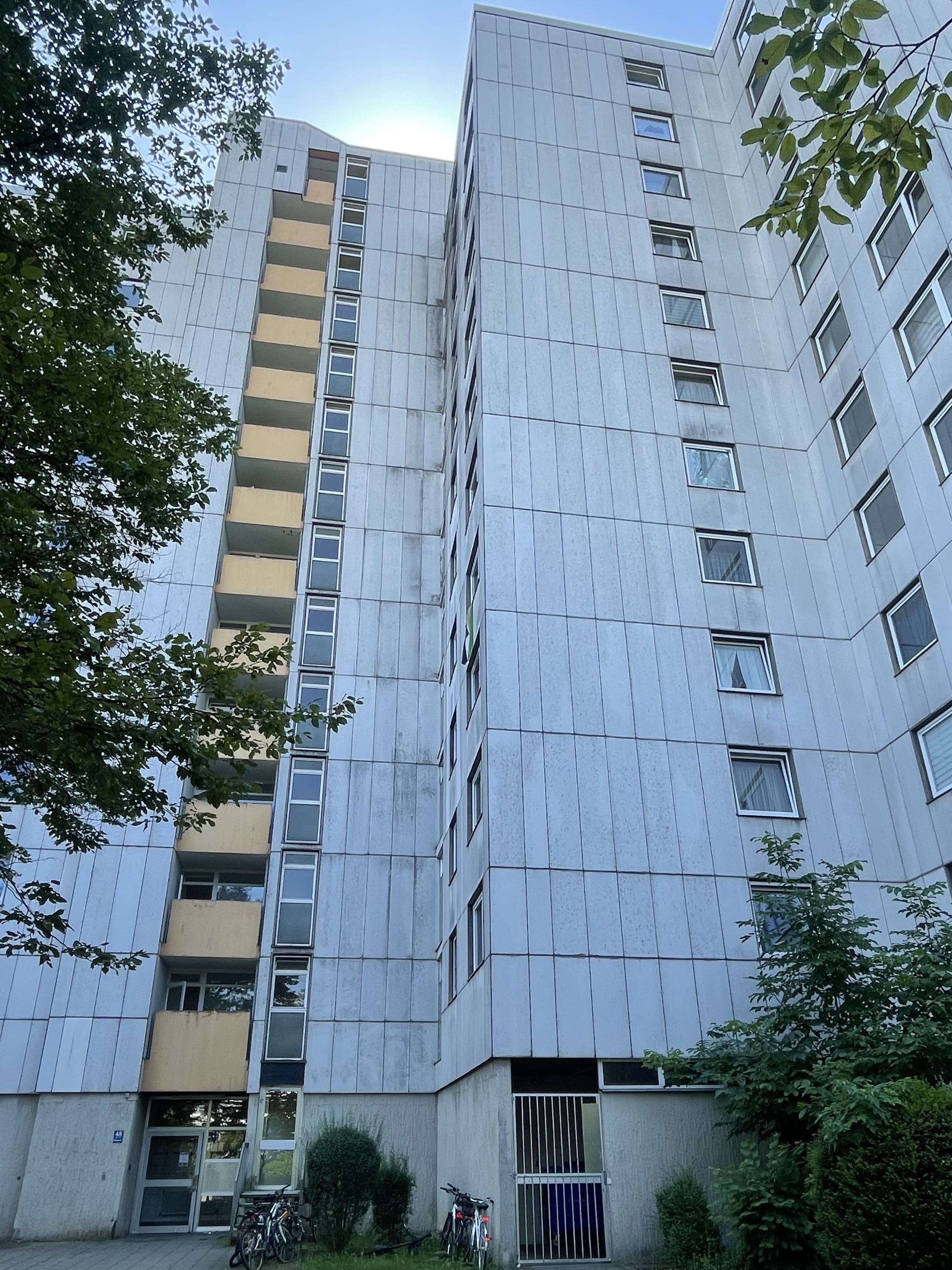}%

    & \includegraphics[width=0.1\linewidth]{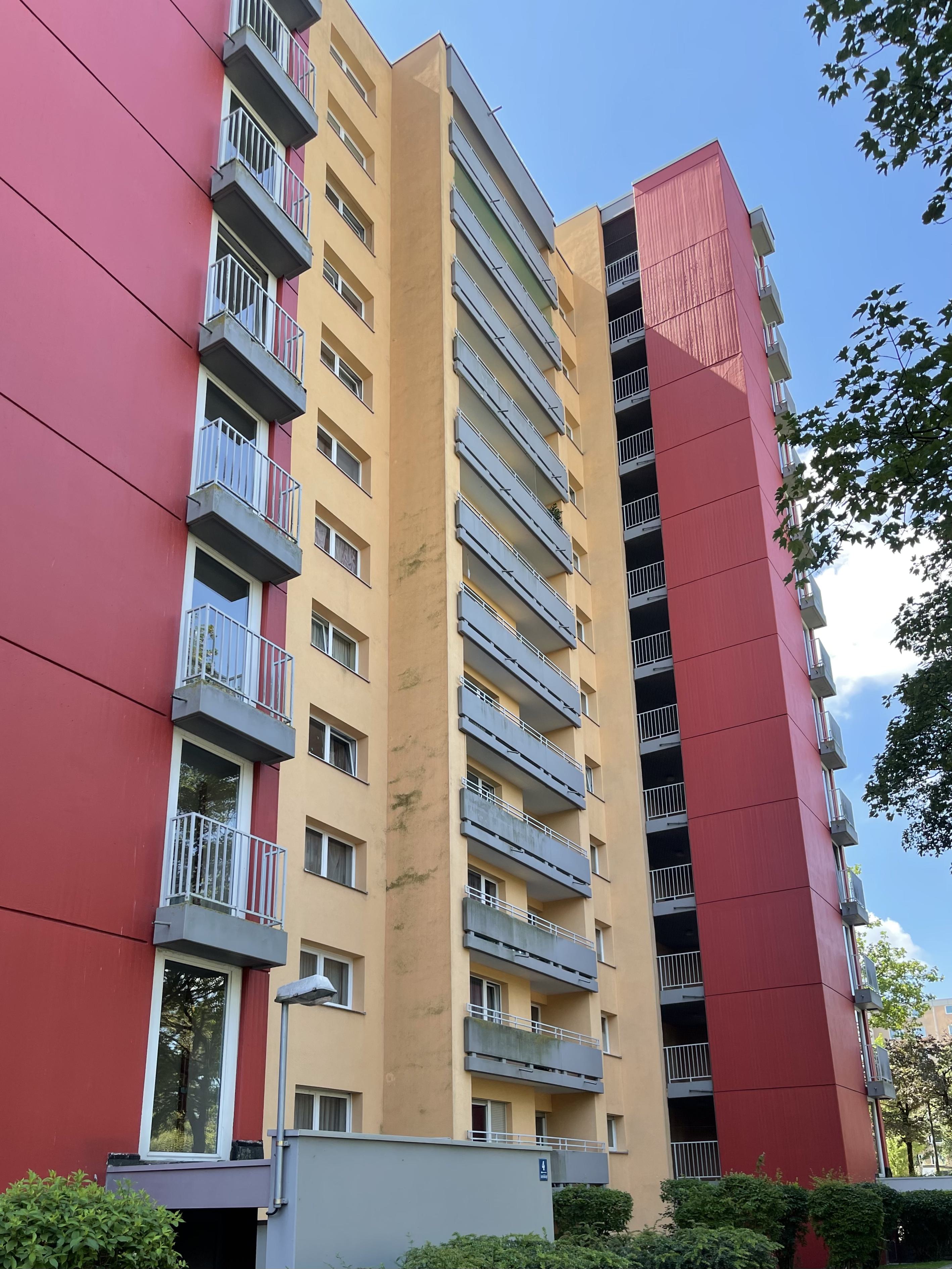}%

    & \includegraphics[width=0.1\linewidth]{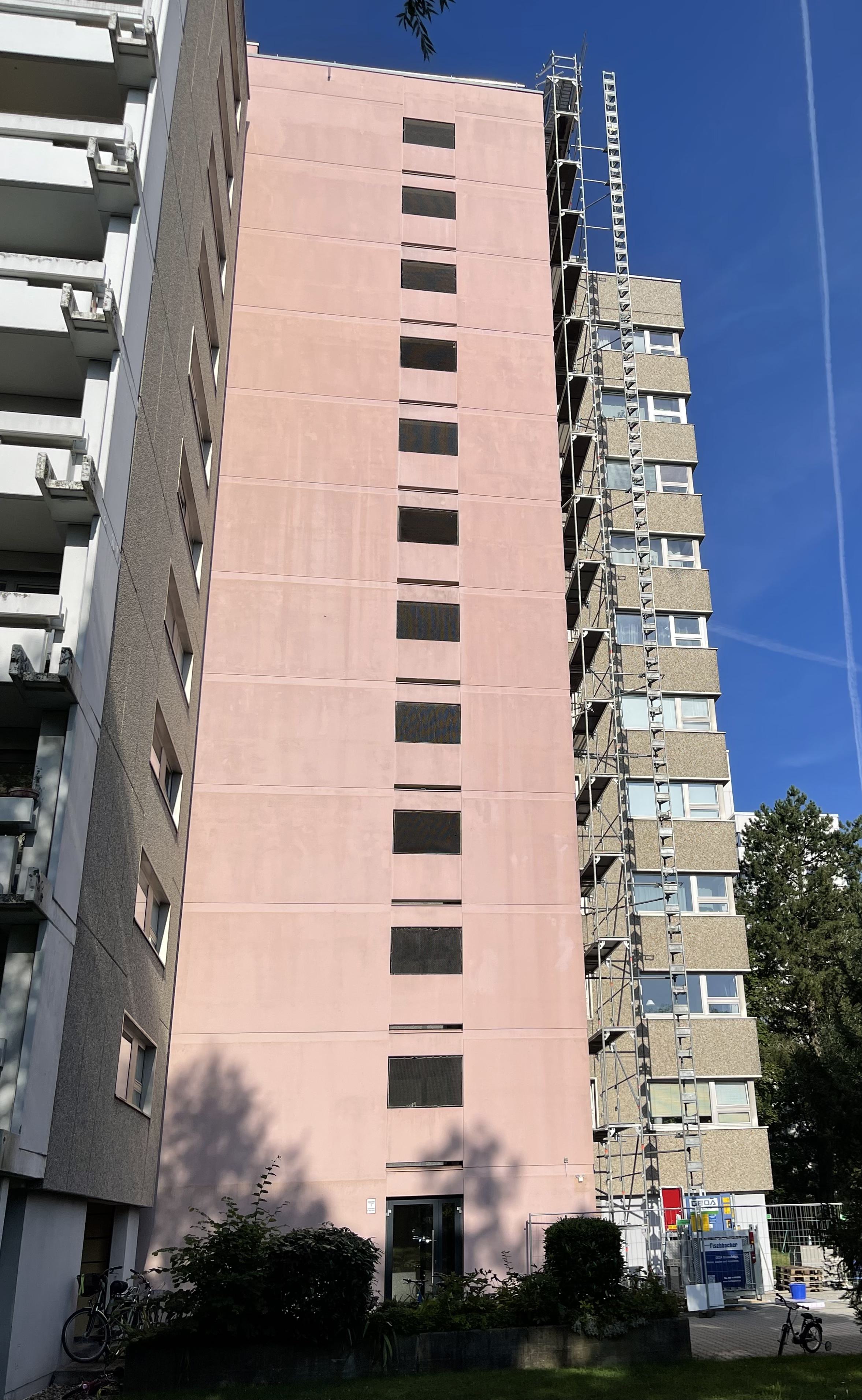}%
\\
    10,08,08&  10,09,09&  10,09,10&  10,10,09&  10,10,10&  11,09,10&  11,11,10&  11,11,11&  11,14,14\\
    \includegraphics[width=0.1\linewidth]{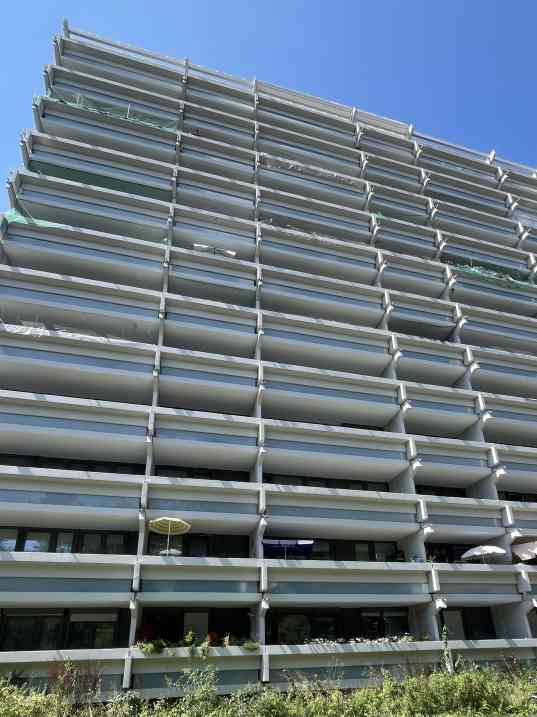}%

    & \includegraphics[width=0.1\linewidth]{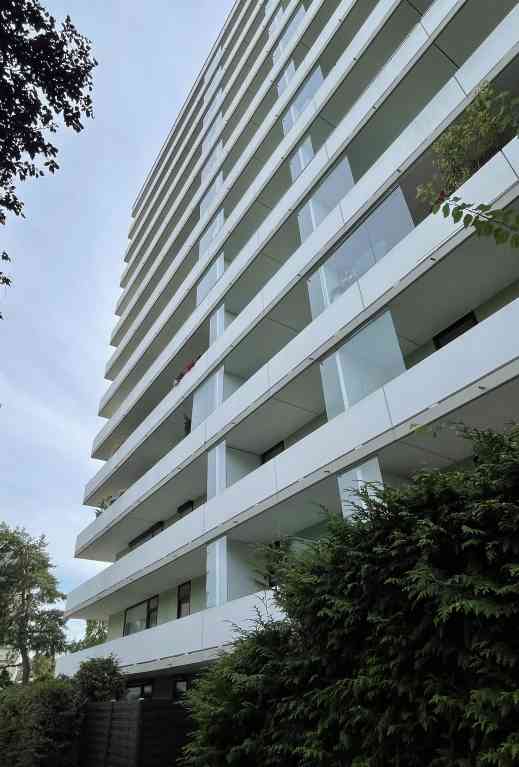}%

    & \includegraphics[width=0.1\linewidth]{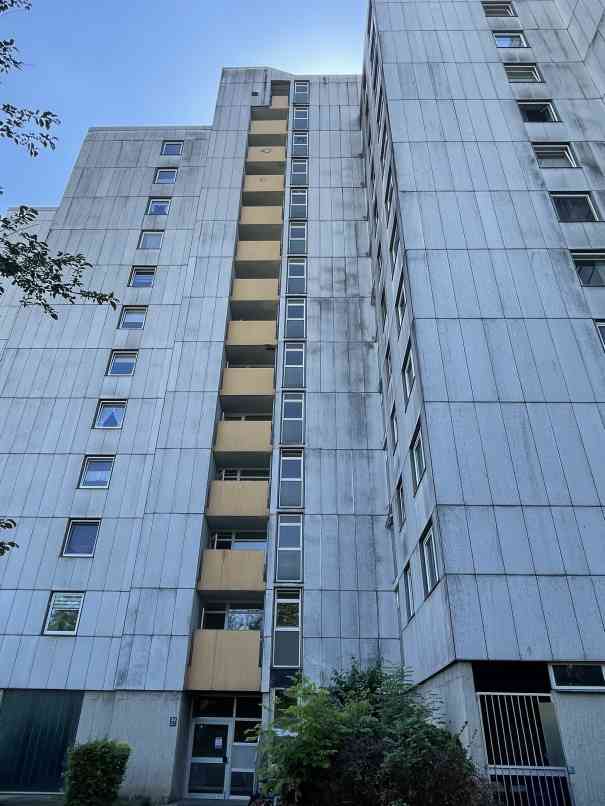}%

    & \includegraphics[width=0.1\linewidth]{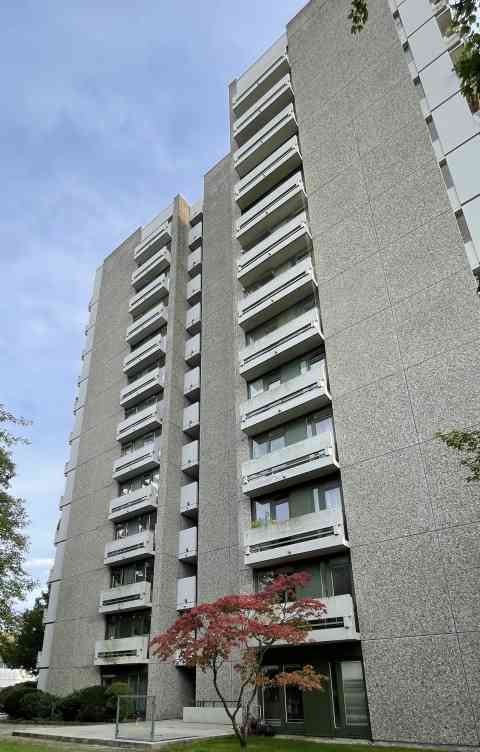}%

    & \includegraphics[width=0.1\linewidth]{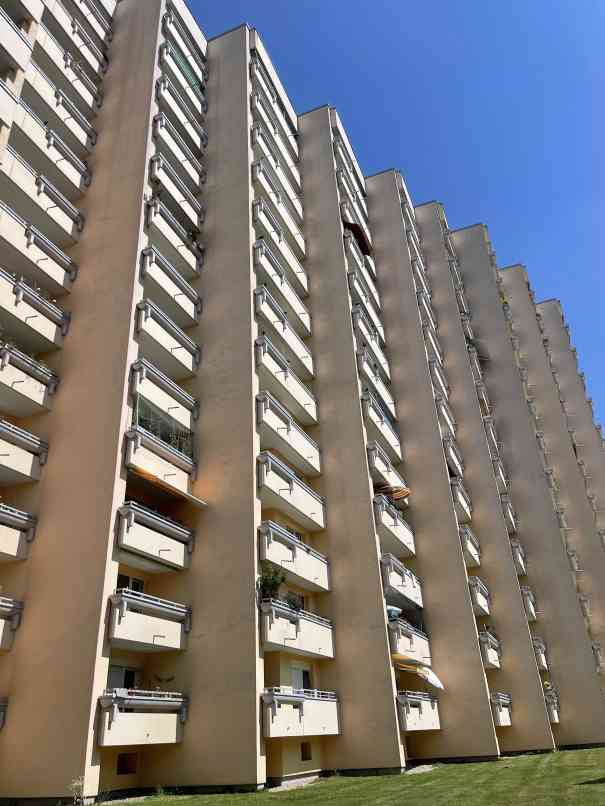}%

    & \includegraphics[width=0.1\linewidth]{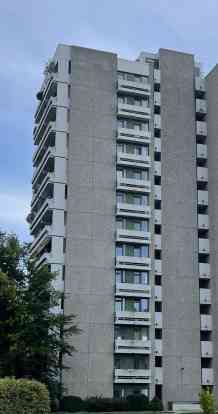}%

    & \includegraphics[width=0.1\linewidth]{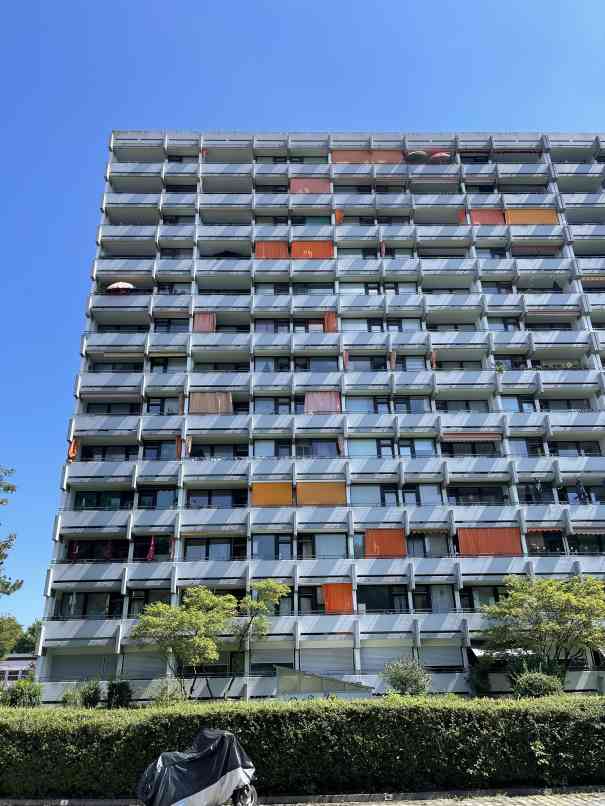}%

    & \includegraphics[width=0.1\linewidth]{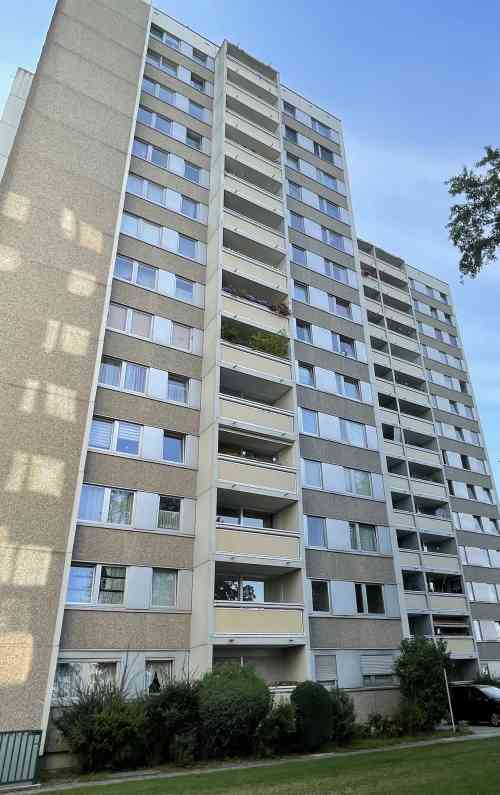}%

    & \includegraphics[width=0.1\linewidth]{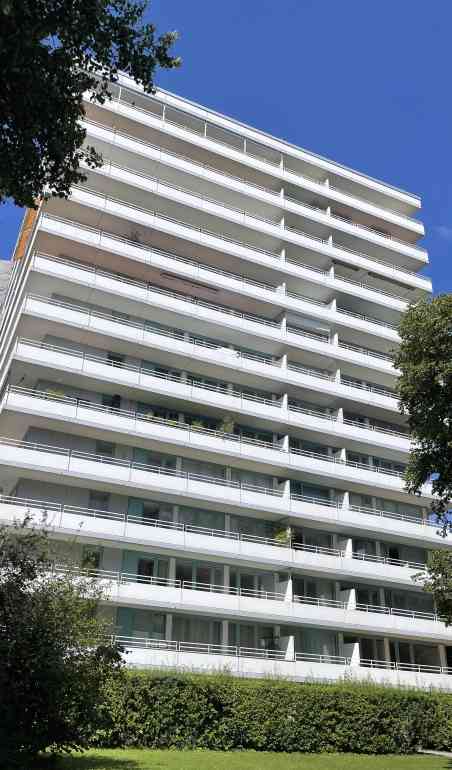}%
\\
    11,15,14&  12,10,10&  12,11,11&  12,12,12&  12,14,13&  12,14,15&  12,15,13&  13,12,12&  13,13,13\\
    \includegraphics[width=0.1\linewidth]{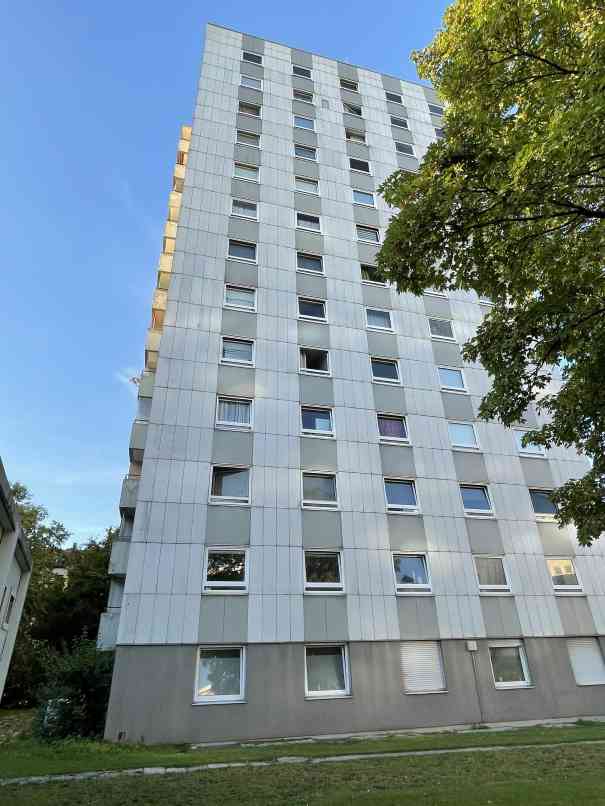}%

    & \includegraphics[width=0.1\linewidth]{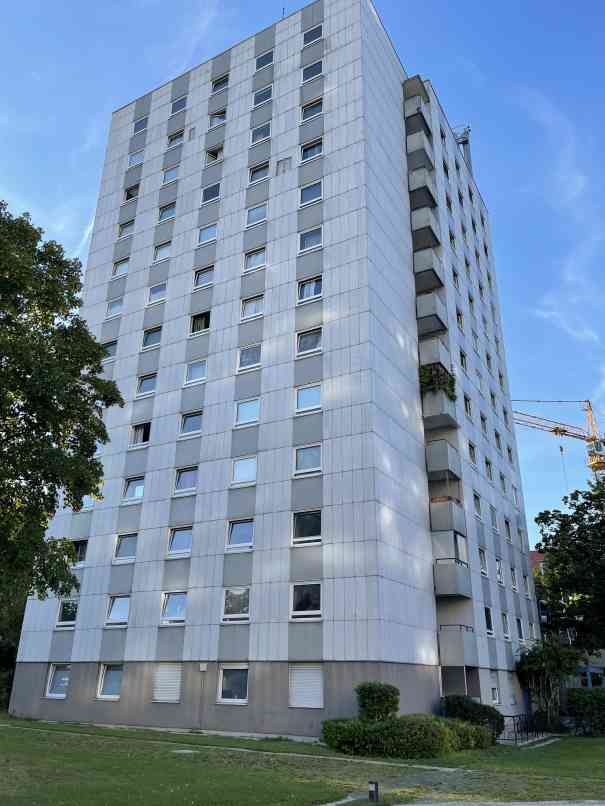}%

    & \includegraphics[width=0.1\linewidth,height=0.13\linewidth]{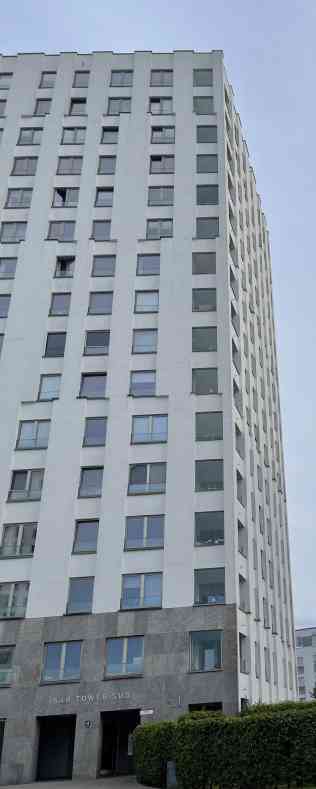}%

    & \includegraphics[width=0.1\linewidth]{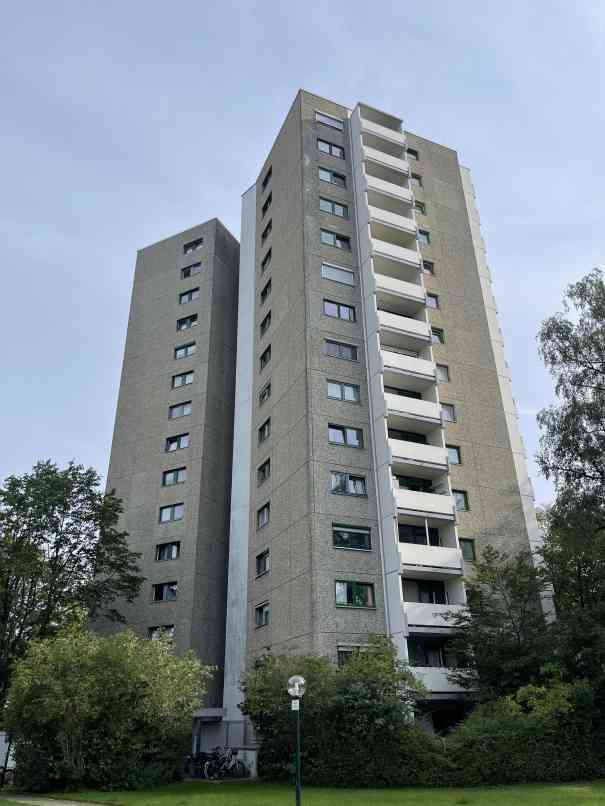}%

    & \includegraphics[width=0.1\linewidth,height=0.13\linewidth]{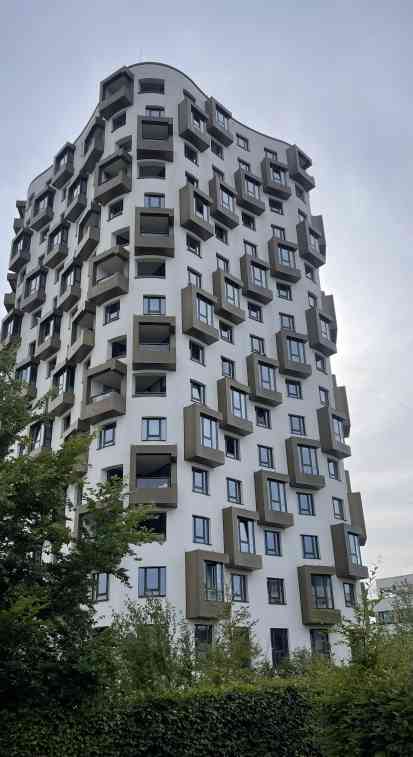}%

    & \includegraphics[width=0.1\linewidth]{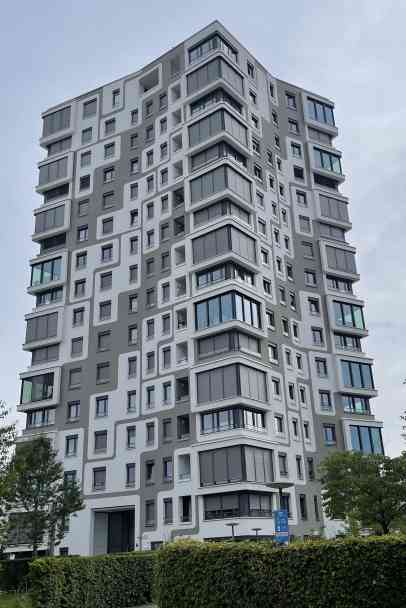}%

    & \includegraphics[width=0.1\linewidth]{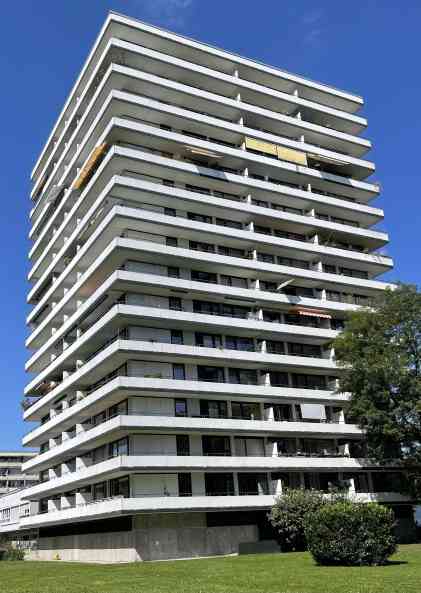}%

    & \includegraphics[width=0.1\linewidth]{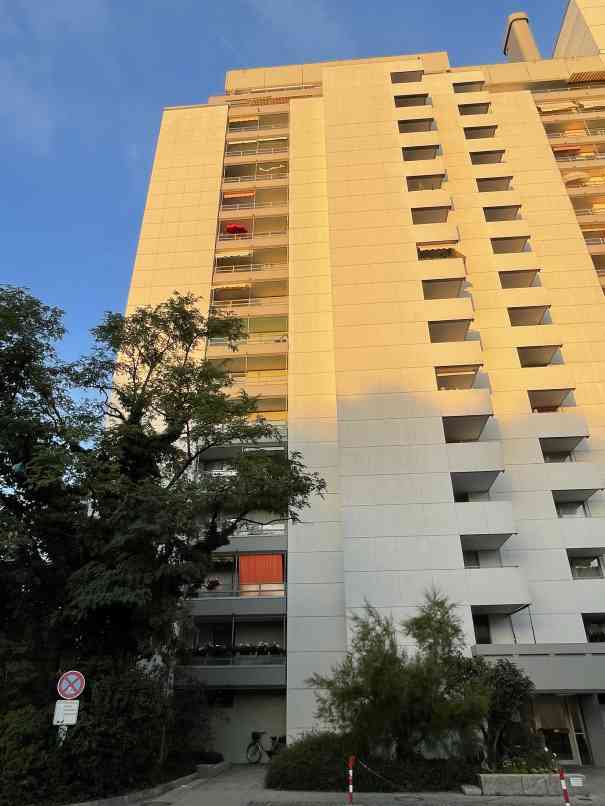}%

    & \includegraphics[width=0.1\linewidth]{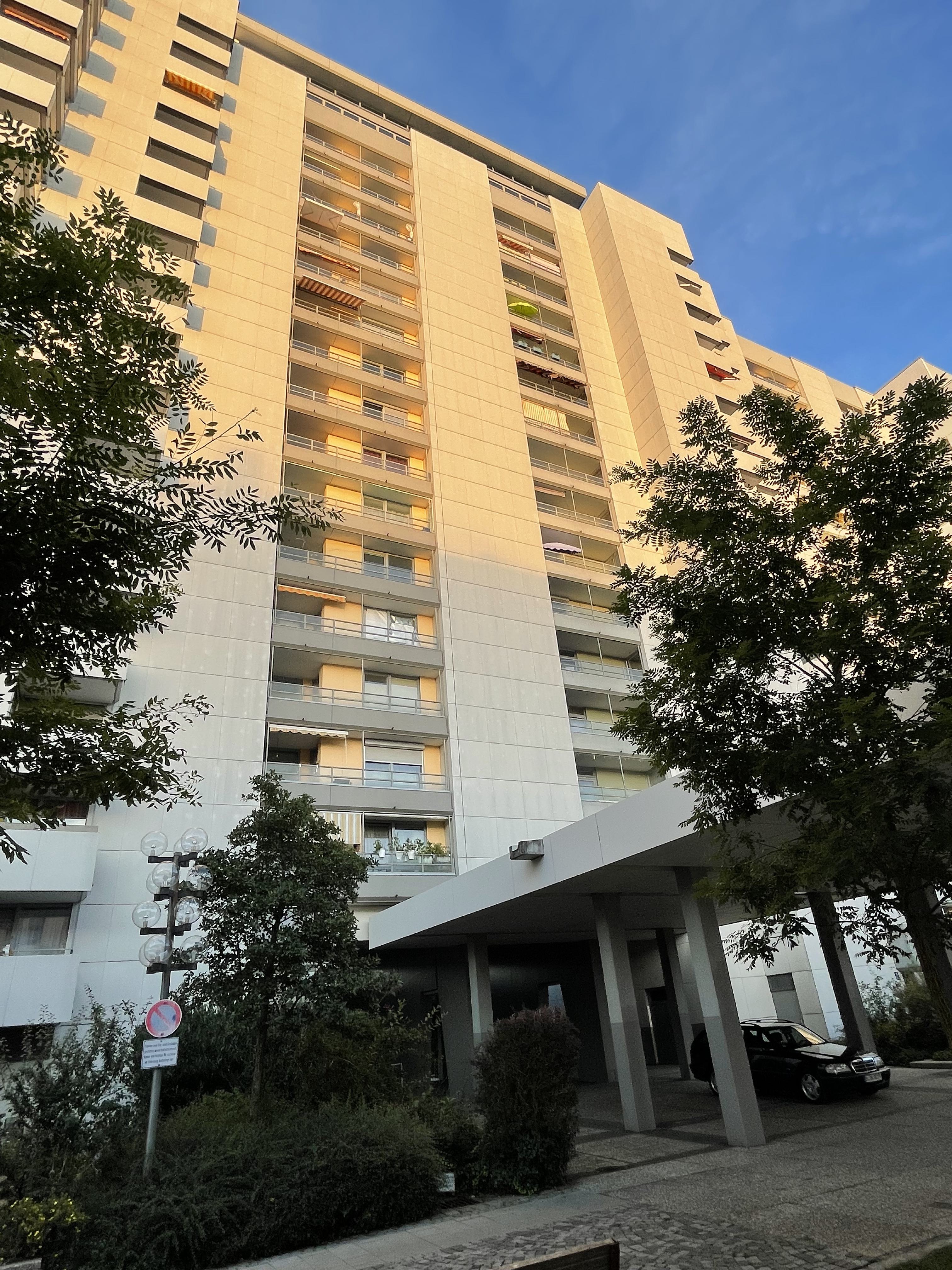}%
\\
    14,13,13&  14,14,11&  14,14,14&  14,16,14&  15,13,12&  15,15,15&  15,15,16&  16,15,15&  17,17,17\\
    \end{tabular}
    \caption{Samples of high-rise buildings (floor number >=8) with ground truth, STL, and MTL predictions. The three numbers under each building image represent the ground truth class (floor number - 1), the STL result, and the MTL result, respectively.}
    \label{fig:grid_images_continue}
\end{figure*}

%% file: sections/06_conlusion.tex
\section{Conclusion}\label{sec:conclude}

This study introduces a scalable and automated framework for estimating building floor numbers using unrestricted, crowdsourced street-level imagery. By leveraging a deep learning approach, the need for manually crafted features is eliminated, ensuring adaptability to diverse architectural styles. 
To support reproducible research, we assembled and released the {Munich Building Floor Dataset}, 6\,800\,+ geo-referenced images drawn from Mapillary and our own high-rise survey, each labelled with a verified floor count. 
The end-to-end pipeline detects façades in raw images, and aligns them with 2-D building footprints.  
A classification-regression network serves as the baseline model, achieving an accuracy of 81.19\% and 97.90\% within a one-floor tolerance, demonstrating the potential of our approach for robust and scalable floor number estimation, particularly in urban environments. 

Future work will extend the approach in two directions: \textit{(i)} enriching the learning objective with complementary cues (e.g.\ roof type or façade segmentation) and \textit{(ii)} expanding the dataset beyond Munich to improve performance and generalizability on this foundation. These steps will further advance large-scale, image-based retrieval of building attributes for urban planning, disaster management, and energy modeling.